\documentclass{article}
\usepackage{graphicx} 
\usepackage[utf8]{inputenc}
\usepackage[]{amsthm} 
\usepackage{caption}
\usepackage[]{amssymb} 
\usepackage[]{amsmath,txfonts}
\usepackage{float}
\usepackage{longtable}
\usepackage{bm}
\usepackage{multirow}
\usepackage{graphicx}
\graphicspath{ {./images/} }
\usepackage[toc,page]{appendix}
\usepackage{lscape}
\usepackage{multirow}
\usepackage{booktabs}
\usepackage{array}    
\usepackage{hyperref}
\hypersetup{
    colorlinks=true, allcolors = blue
    }
\usepackage{authblk}
\usepackage{stackengine}
\usepackage[style=apa]{biblatex}
\title{K-Fold Causal BART for CATE Estimation}
\author[1]{Hugo Gobato Souto}
\affil[1]{\stackunder{{\stackunder{Institute of Mathematics and Computer Sciences at University of São Paulo, Brazil}{Av. Trab. São Carlense 400, 13566-590 São Carlos (SP), Brazil}}}{\stackunder{{hgsouto@usp.br}. {https://orcid.org/0000-0002-7039-0572}}}}
\author[2]{Francisco Louzada Neto}
\affil[2]{\stackunder{{\stackunder{Institute of Mathematics and Computer Sciences at University of São Paulo, Brazil}{Av. Trab. São Carlense, 400, São Carlos, 13566-590, Brazil}}}{\stackunder{{louzada@icmc.usp.br}. {https://orcid.org/0000-0001-7815-9554}}}}
\date{\vspace{-5ex}}

\begin{document}

\maketitle

\begin{abstract}
    This research aims to propose and evaluate a novel model named K-Fold Causal Bayesian Additive Regression Trees (K-Fold Causal BART) for improved estimation of Average Treatment Effects (ATE) and Conditional Average Treatment Effects (CATE). The study employs synthetic and semi-synthetic datasets, including the widely recognized Infant Health and Development Program (IHDP) benchmark dataset, to validate the model's performance. Despite promising results in synthetic scenarios, the IHDP dataset reveals that the proposed model is not state-of-the-art for ATE and CATE estimation. Nonetheless, the research provides several novel insights: 1. The ps-BART model is likely the preferred choice for CATE and ATE estimation due to better generalization compared to the other benchmark models - including the Bayesian Causal Forest (BCF) model, which is considered by many the current best model for CATE estimation, 2. The BCF model's performance deteriorates significantly with increasing treatment effect heterogeneity, while the ps-BART model remains robust, 3. Models tend to be overconfident in CATE uncertainty quantification when treatment effect heterogeneity is low, 4. A second K-Fold method is unnecessary for avoiding overfitting in CATE estimation, as it adds computational costs without improving performance, 5. Detailed analysis reveals the importance of understanding dataset characteristics and using nuanced evaluation methods, 6. The conclusion of \citeauthor{BENCHMARKS2021_2a79ea27} (\citeyear{BENCHMARKS2021_2a79ea27}) that indirect strategies for CATE estimation are superior for the IHDP dataset is contradicted by the results of this research. These findings challenge existing assumptions and suggest directions for future research to enhance causal inference methodologies.
\end{abstract}


\section{Introduction}

\subsection{Origin of Causal Inference and Early Works}
The quest to ascertain causal relationships lies at the heart of empirical research, transcending disciplines to unravel the mechanisms by which interventions alter outcomes \parencite{10.1145/3444944,10.1145/3394486.3406460,Kuang2020, Crown2019, Grimmer2014, Ohlsson2020, Bembom2007,Sobel2000, murnane2010methods}. It is crucial, though, to distinguish between correlation and causation. For instance, observing that ice cream sales and drowning incidents both increase during summer does not imply that ice cream consumption causes drowning, albeit if one performs a simple linear regression in this case, one could conclude this. This correlation is spurious and driven by a common confounder, the warm weather, which influences both variables. Mistaking correlation for causation can lead to false conclusions and ineffective or even harmful interventions. For example, if policymakers incorrectly conclude that ice cream causes drowning, they might implement unnecessary regulations on ice cream sales instead of focusing on water safety during summer months.

Central to the inquiry of determining causal relationships is causal inference, a domain dedicated to elucidating the nature and magnitude of effects engendered by treatments or policies \parencite{10.1145/3444944,Kuang2020}. A foundational concept in causal inference is the idea of potential outcomes. These represent the possible outcomes for an individual or unit under different treatment scenarios. For example, we might consider the potential outcome of a patient if they receive a new drug versus if they do not. The challenge lies in estimating these potential outcomes because, in practice, we only observe one outcome for each individual: either they received the treatment or they did not. Given the causality nature in the human thought (and even other animals thought) \parencite{Vadillo2010AugmentationIC, Vallverd2024, Sloman2015, Penn2007}, one could assume that the causal inference literature started to develop itself already from the beginning of scientific study. 

Nonetheless, the main concepts present in causal inference were formalized only in the 1920s with Neyman's introduction of potential outcomes in randomized experiments (\citeauthor{Neyman1923} (\citeyear{Neyman1923}), translated and reprinted in \citeauthor{Neyman1990} (\citeyear{Neyman1990})), and Fisher's assertion of randomization as the foundation for inference \parencite{Fisher1935}. Despite its seemingly obvious nature, the formal notation for potential outcomes was not introduced until 1923 by Neyman and was initially applied exclusively to randomized experiments, not observational studies. Statisticians used potential outcomes notation in experimental contexts while reverting to realized and observed outcomes for observational data. 

The basic idea that causal effects are comparisons of potential outcomes might seem straightforward today, but its formal application was initially limited to randomized experiments. In these experiments, researchers could control who received the treatment and who did not, making it easier to draw causal conclusions. However, this approach did not immediately extend to observational studies, where treatments are not randomly assigned. Statisticians often used different notations and methodologies for observational data, focusing on the outcomes they could observe directly. It was only in the early seventies, particularly through Donald Rubin's work \parencite{Rubin1974}, that the potential outcomes framework became prominent in observational studies, eventually gaining widespread acceptance over the next 25 years as a standard approach to defining and evaluating causal effects across various settings.

Pivotal concepts within causal inference are the Average Treatment Effect (ATE) and Conditional Average Treatment Effect (CATE), defined as the expected difference in potential outcomes under treatment versus control and the expected difference in potential outcomes under treatment versus control given the unique characteristics of the individual observation. 

\subsection{Problem Statement and Common Notation}
Central to causal inference are the concepts of the Average Treatment Effect (ATE) and the Conditional Average Treatment Effect (CATE). These concepts help quantify the impact of a treatment or intervention. 

To understand these concepts, consider a study evaluating the effect of a new medication on blood pressure. The ATE measures the average effect of the medication across the entire population. It is defined as the expected difference in outcomes between those who receive the treatment and those who do not:
\begin{equation}
    \tau := \mathbb{E}[Y_i | Z_i = 1] - \mathbb{E}[Y_i | Z_i = 0],
\end{equation}
where $\tau$ represents the ATE, $Y_i$ is the outcome for observation $i$, and $Z_i$ is the binary treatment indicator (1 if treated, 0 if not).

However, individuals may respond differently to the treatment based on their specific characteristics, such as age, weight, or genetic factors. This is where the CATE becomes important. The CATE provides an estimate of the treatment effect for subgroups of the population defined by these covariates:
\begin{equation}
    \tau (\mathbf{x_i}) := \mathbb{E}[Y_i | \mathbf{x_i}, Z_i = 1] - \mathbb{E}[Y_i | \mathbf{x_i}, Z_i = 0],
\end{equation}
where $\tau (\mathbf{x_i})$ represents the CATE for observation $i$ given their covariates $\mathbf{x_i}$ (i.e., in the example of a new medication against high blood pressure, $\mathbf{x_i}$ would be the unique characteristics of patient $i$ that potentially have an impact on the treatment effect of the new medication in patient $i$). It worth mentioning that there exist different formulations of ATE and CATE if the treatment $Z$ is continuous instead of binary. Yet, such a problem formulation is outside the scope of this thesis as it encompasses different considerations and approaches from the original and most studied formulation of the treatment $Z$ being binary \parencite{https://doi.org/10.48550/arxiv.2007.09845}.

The relationship between the outcome, treatment, and covariates can be modeled as:
\begin{equation}
    Y_i = f(\mathbf{x_i}, Z_i) + \epsilon_i, \quad \epsilon_i \sim \mathcal{N}(0, \sigma^2),
\end{equation}
where $f(\mathbf{x_i}, Z_i)$ represents the true but unknown relationship, and $\epsilon_i$ is the error term, normally distributed with mean 0 and variance $\sigma^2$. The treatment effect $\tau (\mathbf{x_i})$ can be expressed as:
\begin{equation}
    \tau (\mathbf{x_i}) := f(\mathbf{x_i}, 1) - f(\mathbf{\mathbf{x_i}}, 0),
\end{equation}
isolating the effect of the treatment from other factors.

A critical aspect in causal inference is the estimation of the propensity score, defined as the probability of receiving the treatment given covariates:
\begin{equation}
    \pi(\mathbf{x_i}) = \Pr(Z_i = 1 | \mathbf{x_i}).
\end{equation}
The propensity score helps adjust for confounding in observational studies, where treatment assignment is not randomized \parencite{Pan2018,Shiba2021,Imai2004,Tan2006,Li2012,Fuentes2021}. By balancing covariates between treatment and control groups, the propensity score enables a more accurate estimation of causal effects \parencite{Pan2018,Shiba2021,Imai2004,Tan2006,Li2012,Fuentes2021}.

The estimation of CATE relies on several critical assumptions, notably the Stable Unit Treatment Value Assumption (SUTVA), the Strong Ignorability Assumption, and the Overlap Assumption. SUTVA posits that the potential outcome for any unit is unaffected by the treatment status of others and that there are no multiple versions of each treatment leading to different outcomes. Mathematically, SUTVA can be represented as:
\begin{equation}
    Y_i = Z_iY_i(1) + (1 - Z_i)Y_i(0),
\end{equation}
The Strong Ignorability Assumption ensures that, given covariates $\mathbf{x_i}$, treatment assignment is independent of potential outcomes, facilitating unbiased estimation of treatment effects:
\begin{equation}
    \{Y(0), Y(1)\} \perp Z | \mathbf{x_i}.
\end{equation}

The Overlap Assumption, also known as the Common Support Assumption, requires that for each value of the covariates, there is a positive probability of receiving both the treatment and the control. This ensures that each treated individual can be compared to a similar untreated individual. Mathematically, it can be expressed as:
\begin{equation}
    0 < P(Z_i = 1 | \mathbf{x_i}) < 1 \quad \forall \mathbf{x_i}.
\end{equation}

These assumptions and methodologies form the backbone of causal inference, providing a structured approach to understanding the effects of interventions across diverse settings.

\subsection{Main Approaches to solve the Problem}

The proper estimation of CATE currently has more attention in the causal inference literature than ATE estimation as the former is more complex and with the former, the latter can be easily estimated \parencite{Hahn2020}. As a result, this thesis will mainly handle and focus on the estimation of CATE. Various methods have been developed to estimate CATE, ranging from parametric to nonparametric approaches. Below the main parametric and nonparametric approaches are described. 

\subsubsection*{1. Parametric Manner}

Parametric methods rely on predefined functional forms to model the relationship between covariates, treatment, and outcomes. Common parametric methods include ordinary linear regression and the Lasso method.

\paragraph{a) Ordinary Linear Regression and OLS}
Linear regression is one of the simplest and most widely used methods for estimating treatment effects, especially in Social Sciences studies \parencite{CausalBook}. It assumes that the relationship between the outcome variable \( Y \) and the covariates \( \mathbf{X} \) is linear. The basic linear regression model can be written as:
\begin{equation}
    Y_i = \beta_0 + \beta_1 Z_i + \beta_2 x_{i1} + \beta_3 x_{i2} + \ldots + \beta_k x_{ik} + \epsilon_i,
\end{equation}
where, \( \beta_0, \beta_1, \ldots, \beta_k \) are the coefficients to be estimated. It is worth mentioning that \( \mathbf{X} \) could actually be nonlinear transformations of the real covariates, possibly giving even more flexibility and power to the model if the transformations are appropriate \parencite{CausalBook}.

In this case, the coefficient \( \beta_1 \) represents the average treatment effect (ATE). Nevertheless, to estimate the CATE, we can introduce interaction terms between the treatment and covariates:
\begin{equation}
    Y_i = \beta_0 + \beta_1 Z_i + \sum_{j=1}^k \beta_j x_{ij} + \sum_{j=1}^k \gamma_j (Z_i \cdot x_{ij}) + \epsilon_i.
\end{equation}

In this extended model:
\begin{itemize}
    \item \( \gamma_j \) are the coefficients for the interaction terms between treatment \( Z_i \) and covariates \( x_{ij} \).
\end{itemize}

The CATE for an individual with covariates \( x_i \) is given by:
\begin{equation}
    \tau(x_i) = \beta_1 + \sum_{j=1}^k \gamma_j x_{ij}.
\end{equation}

Such a model allows the treatment effect to vary with the covariates, providing a more nuanced estimate of the treatment effect for different subpopulations \parencite{CausalBook}.

\paragraph{b) Lasso Method}
The Lasso (Least Absolute Shrinkage and Selection Operator) method \parencite{Tibshirani1996} is an extension of linear regression that addresses some of the limitations of OLS, particularly in high-dimensional settings. Lasso introduces a regularization term to the OLS objective function, which penalizes the absolute size of the regression coefficients. This penalty encourages sparsity, effectively selecting a subset of covariates that have the most significant impact on the outcome \parencite{Tibshirani1996}.

The Lasso objective function is given by:
\begin{equation}
    \hat{\bm{\beta}} = \arg\min_{\bm{\beta}} \left\{ \sum_{i=1}^n (Y_i - \beta_0 - \beta_1 Z_i - \sum_{j=1}^k \beta_j x_{ij} - \sum_{j=1}^k \gamma_j (Z_i \cdot x_{ij}))^2 + \lambda \sum_{j=1}^k |\beta_j| \right\},
\end{equation}
where \( \lambda \) is a regularization parameter that controls the trade-off between fitting the data well and keeping the model coefficients small.

The Lasso estimates are obtained by solving this optimization problem, which can be done using algorithms such as coordinate descent \parencite{Tibshirani1996}. The regularization parameter \( \lambda \) is typically selected using cross-validation, where the model is trained on a subset of the data and validated on the remaining data to find the value of \( \lambda \) that minimizes the prediction error \parencite{Tibshirani1996}.

The Lasso method provides several advantages over OLS:
\begin{itemize}
    \item \textbf{Variable Selection}: By shrinking some coefficients to zero, Lasso effectively selects a subset of relevant covariates, which is useful in high-dimensional settings where many covariates may be irrelevant.
    \item \textbf{Improved Prediction Accuracy}: The regularization term helps prevent overfitting, leading to more robust and generalizable models.
    \item \textbf{Interpretability}: Sparse models are easier to interpret, as they involve fewer covariates, making it clearer which covariates are most influential in predicting the outcome.
\end{itemize}

\subsubsection*{2. Nonparametric Manners}

Nonparametric methods do not assume a specific functional form for the relationship between covariates, treatment, and outcomes. Instead, they rely on data-driven techniques to estimate CATE. As a result, they usually have a more powerful predictive and inference power as it is quite a challenging to correctly parametrize a model for a certain problem \parencite{pmlr-v130-curth21a}. Key nonparametric methods for CATE estimation include S-Learner, T-Learner, DR-Learner, R-Learner, and X-Learner \parencite{pmlr-v130-curth21a, CausalBook}. These are described below.

\paragraph{a) Single (S)-Learner}
The Single (S)-Learner is a straightforward and intuitive approach to estimate CATE using machine learning models \parencite{Knzel2019,CausalBook}. The fundamental idea behind the S-Learner is to train a single model that directly incorporates both the treatment variable and the covariates to predict the outcome. This approach leverages the power of machine learning to capture complex relationships between the covariates, treatment, and outcome.

The S-Learner approach involves training a single predictive model \( f \) to estimate the potential outcomes for each individual. The model takes as input the covariates \( \mathbf{x_i} \) and the treatment indicator \( Z_i \), and outputs the predicted outcome \( \hat{Y}_i \). The general form of the model can be expressed as:
\begin{equation}
    \hat{Y}_i = f(\mathbf{x_i}, Z_i),
\end{equation}
where \( \hat{Y}_i \) is the predicted outcome for observation \( i \), \( \mathbf{x_i} \) represents the covariates, and \( Z_i \) is the binary treatment indicator (1 if treated, 0 if not).

To estimate the potential outcomes \( Y_i(1) \) and \( Y_i(0) \) for each individual, the S-Learner uses the trained model to make two predictions:
\begin{align}
    \hat{Y}_i(1) &= f(x_i, 1), \\
    \hat{Y}_i(0) &= f(x_i, 0).
\end{align}
Here, \( \hat{Y}_i(1) \) is the predicted outcome when the treatment is applied, and \( \hat{Y}_i(0) \) is the predicted outcome when the treatment is not applied.

Once the potential outcomes are estimated, the CATE for an individual with covariates \( \mathbf{x_i} \) can be computed as the difference between the predicted potential outcomes:
\begin{equation}
    \hat{\tau}(\mathbf{x_i}) = \hat{Y}_i(1) - \hat{Y}_i(0).
\end{equation}
This difference represents the estimated effect of the treatment on the outcome for an individual with characteristics \( \mathbf{x_i} \).

The training process for the S-Learner involves fitting the model \( f \) to the observed data \((\mathbf{x_i}, Z_i, Y_i)\). This can be done using various machine learning algorithms, such as linear regression, decision trees, or neural networks \parencite{Knzel2019,CausalBook}. The choice of algorithm depends on the complexity of the relationship between the covariates, treatment, and outcome, as well as the available computational resources \parencite{Knzel2019,CausalBook}.

The objective is to minimize the prediction error, which is typically measured by the Mean Squared Error (MSE) \parencite{Knzel2019,CausalBook}:
\begin{equation}
    \text{MSE} = \frac{1}{n} \sum_{i=1}^n (Y_i - f(\mathbf{x_i}, Z_i))^2,
\end{equation}
where \( n \) is the number of observations, \( Y_i \) is the observed outcome, and \( f(\mathbf{x_i}, Z_i) \) is the predicted outcome.

Nevertheless S-Learner has a main limitation: when estimating a regression model that predicts the outcome based on the treatment and covariates, the treatment variable can be excessively regularized. This issue is particularly prominent in scenarios where the treatment effect is small. In such cases, many machine learning algorithms may tend to reduce the treatment effect towards zero, giving more importance to other informative covariates in the model \parencite{Knzel2019,CausalBook}. To solve this issue, the T-Learner approach can be used \parencite{Knzel2019,CausalBook}.

\paragraph*{b) Two (T)-Learner}

The Two (T)-Learner is a more sophisticated approach for estimating the CATE by employing two separate models for the treatment and control groups. This method allows for a more nuanced analysis of treatment effects by explicitly modeling the outcomes for treated and untreated individuals separately \parencite{Knzel2019,CausalBook}. The T-Learner approach is particularly useful for capturing complex interactions between treatment effects and covariates.

In the Two (T)-Learner approach, two distinct models are trained to predict the potential outcomes for the treated and control groups. Let \( f_T(\mathbf{x_i}) \) denote the model for the treated group and \( f_C(\mathbf{x_i}) \) denote the model for the control group. The models are formulated as:
    \begin{align}
        \hat{Y}_i(1) &= f_T(\mathbf{x_i}), \\
        \hat{Y}_i(0) &= f_C(\mathbf{x_i}),
    \end{align}
where \( \hat{Y}_i(1) \) represents the predicted outcome for the treated group and \( \hat{Y}_i(0) \) represents the predicted outcome for the control group. The CATE is estimated as the difference between the predicted outcomes for the treatment and control conditions:
    \begin{equation}
        \hat{\tau}(\mathbf{x_i}) = \hat{Y}_i(1) - \hat{Y}_i(0).
    \end{equation}

The training process for the T-Learner is very similar to the S-Learner, but now two models are regressed and not only one. Nonetheless, both the S-Learner and T-Learner depend heavily on the accurate modeling of outcomes. In situations where the conditional counterfactual outcomes \( \mathbb{E}[Y | Z = 1, \mathbf{x_i}] \) are complex and difficult to model, yet the CATE function \( \tau(\mathbf{x_i}) \) is relatively straightforward, these meta-learners may encounter significant estimation errors in the functions \( f, f_T, \) and \( f_C \). To address this, the DR-Learner can be used.

\paragraph*{c) Doubly Robust (DR)-Learner}

The key feature of the DR-Learner is that it combines both outcome regression and propensity score weighting to produce estimates that are robust to misspecification in either the outcome model or the treatment model, but not both \parencite{https://doi.org/10.48550/arxiv.2004.14497,CausalBook}. This doubly robust property ensures more reliable and consistent estimates of CATE, even when one of the models is misspecified \parencite{https://doi.org/10.48550/arxiv.2004.14497,CausalBook}.

The DR-Learner leverages two primary components: an outcome model and a propensity score model. The outcome model predicts the outcome given covariates and treatment, while the propensity score model estimates the probability of receiving the treatment given covariates. The combination of these two models leads to the doubly robust estimation. In mathematical terms:

1. Estimating the Outcome Models:
    \begin{align}
        \hat{Y}_1(\mathbf{x_i}) &= f_1(\mathbf{x_i}), \\
        \hat{Y}_0(\mathbf{x_i}) &= f_0(\mathbf{x_i}),
    \end{align}
    where \( f_1 \) and \( f_0 \) are models trained on the treated and control groups, respectively.

2. Estimating the Propensity Score Model:
    \begin{equation}
        \pi(\mathbf{x_i}) = \Pr(Z_i = 1 | \mathbf{x_i}),
    \end{equation}
    where \( \pi(\mathbf{x_i}) \) is the estimated probability that an individual with covariates \( \mathbf{x_i} \) receives the treatment.

3. Computing the Doubly Robust Estimate:
    The doubly robust estimator for CATE combines the outcome regression and propensity score models:
    \begin{equation}
        \hat{\tau}_{DR}(\mathbf{x_i}) = \left( \frac{Z_i - \pi(\mathbf{x_i})}{\pi(\mathbf{x_i})(1 - \pi(\mathbf{x_i}))} \right) \left( Y_i - \hat{Y}_{Z_i}(\mathbf{x_i}) \right) + \left( \hat{Y}_1(\mathbf{x_i}) - \hat{Y}_0(\mathbf{x_i}) \right).
    \end{equation}

In this equation, the first term adjusts for any imbalance in the covariates using the propensity score, while the second term provides the direct estimate from the outcome models. This combination ensures that the estimator remains consistent if either the outcome model or the propensity score model is correctly specified \parencite{https://doi.org/10.48550/arxiv.2004.14497,CausalBook}.

\paragraph*{d) Residual (R)-Learner}
Even if we assume that the true CATE model belongs to a simple function space and we know the true nuisance parameters (i.e., $f_1, f_0,$ and $\pi(\mathbf{x_i})$), the labels used in the final stage of the DR-Learner might still be large due to the division by the propensity score \parencite{https://doi.org/10.48550/arxiv.1712.04912,CausalBook}. In cases where the Overlap Assumption is nearly violated in certain regions of the covariate space, the regression labels \( \hat{\tau}_{DR}(\mathbf{x_i}) \) can become very large in absolute magnitude, leading to high-variance estimates, as we would almost divide $ Z_i - \pi(\mathbf{x_i})$ by zero \parencite{https://doi.org/10.48550/arxiv.1712.04912,CausalBook}. To bypass this problem, the Residual (R)-Learner can be used \parencite{https://doi.org/10.48550/arxiv.1712.04912,CausalBook}.

Essentially, the R-Learner approach estimates CATE using the residual-on-residual method by solving the following minimization problem:
\begin{equation}
    \min_\tau \mathbb{E}(\hat{Y}-\tau(\mathbf{X}) \hat{Z})^2
\end{equation}
where $\hat{Y}=Y-\mathbb{E}(Y\mid \mathbf{x_i})$ and $\hat{Z}=Z-\mathbb{E}(Z\mid \mathbf{x_i})$. Needless to say, $\mathbb{E}(Y\mid \mathbf{x_i})$ and $\mathbb{E}(Z\mid \mathbf{x_i})$ would be estimated by using a certain nonparametric model.

\paragraph*{e) Cross (X)-Learner}

Lastly, the Cross (X)-Learner integrates propensity scores to enhance outcome modeling in a qualitatively different manner compared to the DR- or R-learners, focusing on accuracy and covariate shift rather than reducing sensitivity to errors in the nuisance models. It starts from the premise that the high-dimensional CATE ($ \tau(\mathbf{X})$) is identical when measured on both treated and control groups \parencite{CausalBook}. This means that the Conditional Average Treatment Effect on the Treated (CATT) is equal to the Conditional Average Treatment Effect on the Control (CATC), unlike the average treatment effect, which can differ due to varying distributions of \( \mathbf{X} \) in treatment and control groups \parencite{CausalBook}. Hence, we can identify the CATT and CATC as:
\begin{align}
    \tau^T(\mathbf{X})=\mathbb{E}[Y - \mathbb{E}[Y \mid \mathbf{X}, D = Z] \mid \mathbf{X}, Z = 1] \\
    \tau^C(\mathbf{X})=\mathbb{E}[\mathbb{E}[Y \mid \mathbf{X}, Z = 1] - Y \mid \mathbf{X}, Z = 0]
\end{align}
This method provides two distinct ways to identify $ \tau(\mathbf{X})$, and any convex combination of these solutions also serves as a valid identification strategy for CATE \parencite{CausalBook}. This approach allows us to circumvent the need to accurately model both response functions across all regions of the covariate space, which is necessary for the other Learners. This can be particularly advantageous if the CATE is simpler to learn compared to the mean counterfactual response model.

If the challenge lies in modeling the mean counterfactual response under a treatment rather than the treatment effect, the following strategy can be employed: in regions of the covariate space, denoted as $ \mathbf{S}$ where more control data is available (i.e., \( \pi(\mathbf{S}) \) is small), we can use the CATT approach, which only requires estimating the mean counterfactual response under control, \( \mathbb{E}[Y | \mathbf{X}, Z = 0] \), not under treatment. Although the effect function must still be learned using the treated data, the simplicity of the effect function makes this less problematic even with fewer treated observations.

Similarly, in regions of the covariate space with more treated data (i.e., \( \pi(\mathbf{S}) \) is large), the CATC approach can be applied, which only necessitates estimating the mean counterfactual response under treatment, \( \mathbb{E}[Y | \mathbf{X}, Z = 1] \), not under control. This motivates the use of the following convex combination as the final identification formula for CATE:
\begin{equation}
    \tau(\mathbf{X}) = \tau^T(\mathbf{X}) (1 - \pi(\mathbf{X})) + \tau^C(\mathbf{X}) \pi(\mathbf{X}),
\end{equation}
where \( \tau^T(\mathbf{X}) \) and \( \tau^C(\mathbf{X}) \) are the estimated CATT and CATC, respectively.

\subsection{Aim of this Thesis and Thesis Structure}
As the reader may already have guessed, the recent past literature in the CATE estimation is mainly composed of using novel nonparametric statistical models for a certain Meta Learner (S-Learner, T-Learner, etc.) or tweaking the Meta Learners to improve their performance (or even both at times). Although for some, especially Reviewer 2 (joke intended), such research type can be seem as not "meaningful" scientific development as commonly no novel theoretical insights come from such studies, they do have an important impact on Health and Social Sciences and even Industry since people can use the new most powerful model available for their applied studies to then discover novel things that lead to "meaningful" scientific development. This is the case since the more powerful the used model in a Health or Social Sciences study is regarding CATE estimation, the less likely it is that the estimated CATE and ATE greatly vary from the true CATE and ATE of the considered problem; hence, the closer the conclusions drawn from the estimated CATE and ATE are from the truth. This thesis aims to contribute to the existing literature by proposing a novel approach to estimate CATE by using a novel nonparametric statistical model and tweaking the Meta Learners a bit. The novel approach is coined K-Fold Causal BART and is explained in detail later in this thesis.

This research contributes to the current CATE estimation literature by:

\begin{itemize}
    \item Demonstrating that the ps-BART model generalizes better than the Bayesian Causal Forest (BCF) model \parencite{Hahn2020} across both synthetic and real-world datasets, suggesting it as a preferred model for CATE and ATE estimation.
    \item Revealing that the BCF model's performance deteriorates significantly with increasing treatment effect heterogeneity, whereas the ps-BART model maintains a greater robustness, indicating its suitability in high heterogeneity contexts.
    \item Highlighting the tendency of models to be overconfident in CATE uncertainty quantification when treatment effect heterogeneity is low, suggesting the need for future research to address this issue.
    \item Emphasizing the importance of understanding dataset characteristics and employing nuanced analysis methods to fully grasp model performance, guiding better model selection and evaluation.
    \item Contradicting previous conclusions of \citeauthor{BENCHMARKS2021_2a79ea27} (\citeyear{BENCHMARKS2021_2a79ea27}) that indirect strategies for CATE estimation are superior for the Infant Health and Development Program (IHDP) dataset.
\end{itemize}

The remaining of this thesis is organized as: Section \ref{Literature Review} contains a detailed literature review in the recent past development the literature of CATE estimation. Section \ref{K-Fold Causal BART} details the novel approach presented in this thesis while Section \ref{Research Design} explains the research design utilized to test the K-Fold Causal BART model against benchmark models. Section \ref{Results} presents and discusses the results of the relative performance of the K-Fold Causal BART model for both fully synthetic and semi-synthetic datasets. Lastly, Section \ref{Conclusion and Novel Insights} presents the main conclusions drawn from the results of this thesis and novel insights inferred from them, while Section \ref{Research Limitations and Reflections} finishes the thesis by exploring the limitations of this thesis (and consequently its results). 

\section{Literature Review}\label{Literature Review}

\subsection{Main Models}

\subsubsection{Linear Regression and LASSO Method}
Given the fact that linear regression requires a correct parametrization of the faced problem to properly estimate CATE and ATE (albeit the use of the LASSO method decreases this issue), research focused on the use of linear regression for CATE estimation has lost its popularity in the literature in comparison with nonparametric models given their flexibility, power, and recent greater accessibility due to a higher computational power access by researchers \parencite{https://doi.org/10.48550/arxiv.1806.01888,CausalBook}. Yet, research about the use of linear regression for CATE estimation still exists and will continue to do so as for linear problems, the use of linear regression still yields more accurate CATE estimation than nonparametric models \parencite{Hahn2020,https://doi.org/10.48550/arxiv.1806.01888}.

For example, \citeauthor{Cattaneo2018} (\citeyear{Cattaneo2018}) contributes significantly to this area by addressing the limitations of traditional linear regression models in high-dimensional settings. Their work focuses on providing inference methods that can accommodate many covariates and heteroskedasticity for CATE estimation, conditions often encountered in empirical research. Specifically, they utilize high-dimensional approximations where the number of covariates can grow with the sample size, which is a common scenario in modern datasets. \citeauthor{Cattaneo2018} (\citeyear{Cattaneo2018}) identify that the usual versions of Eicker-White heteroskedasticity consistent standard error estimators are inconsistent under these high-dimensional asymptotics. To address this, they propose a new heteroskedasticity consistent standard error formula that is both automatic and robust. This new formula can handle conditional heteroskedasticity of unknown form and the inclusion of many covariates, making it highly relevant for practitioners dealing with complex datasets.

\citeauthor{Zhang2013} (\citeyear{Zhang2013}), on the other hand, propose methodologies for statistical inference of low-dimensional parameters in high-dimensional data settings. Their work focuses on constructing confidence intervals for individual coefficients and linear combinations of coefficients in a linear regression model. \citeauthor{Zhang2013} (\citeyear{Zhang2013}) provide sufficient conditions for the asymptotic normality of the proposed estimators along with a consistent estimator for their finite-dimensional covariance matrices. These conditions accommodate scenarios where the number of variables exceeds the sample size and where many small non-zero coefficients are present. Their methods apply to both the interval estimation of a preconceived regression coefficient or contrast and the simultaneous interval estimation of many regression coefficients. Anew, situations found in empirical research where CATE estimation is important \parencite{Zhang2013,Belloni2014}.

\citeauthor{https://doi.org/10.48550/arxiv.1806.05081} (\citeyear{https://doi.org/10.48550/arxiv.1806.05081}) extend the application of high-dimensional regression models by considering systems of regression equations with temporal and cross-sectional dependencies in covariates and error processes, something that can be present in various situations of empirical research \parencite{CausalBook}. They develop methodologies for estimation and inference, focusing on constructing robust variable selection procedures using LASSO. To address the multiplicity of equations and dependencies in the data, they propose a block multiplier bootstrap procedure for selecting the overall penalty level, ensuring the validity of their methods under high-dimensional settings. The authors derive oracle properties with a jointly selected tuning parameter and provide debiased simultaneous inference on multiple target parameters within the system. The bootstrap consistency of their test procedures is established using a general Bahadur representation for Z-estimators with dependent data. Finally, the authors apply their methodology to quantify spillover effects of textual sentiment indices in financial markets and to test the connectedness among different sectors. This empirical application highlights the relevance and utility of their methods in real-world high-dimensional data settings, particularly in finance where temporal and cross-sectional dependencies are prevalent.

Moving to the study of \citeauthor{https://doi.org/10.48550/arxiv.2403.03240} (\citeyear{https://doi.org/10.48550/arxiv.2403.03240}), \citeauthor{https://doi.org/10.48550/arxiv.2403.03240} (\citeyear{https://doi.org/10.48550/arxiv.2403.03240}) delves into the estimation and statistical inference of CATE using a new technique coined Triple Lasso. The study data-generating process assumes linear models for outcomes associated with binary treatments, defining the CATE as the difference between the expected outcomes of these linear models. The study allows these linear models to be high-dimensional but emphasizes consistent estimation and statistical inference for CATE without assuming direct sparsity. Instead, the author considers sparsity in the difference between the linear models. The author employs a doubly robust estimator to approximate this difference and subsequently regress the difference on covariates using Lasso regularization. Although this regression estimator is consistent for CATE, the \citeauthor{https://doi.org/10.48550/arxiv.2403.03240} (\citeyear{https://doi.org/10.48550/arxiv.2403.03240}) further reduces bias using techniques from double/debiased machine learning (DML) and debiased Lasso, achieving \(\sqrt{n}\)-consistency and constructing confidence intervals. \citeauthor{https://doi.org/10.48550/arxiv.2403.03240} (\citeyear{https://doi.org/10.48550/arxiv.2403.03240}) refers to this debiased estimator as the triple/debiased Lasso (TDL), applying both DML and debiased Lasso techniques.

Lastly, \citeauthor{https://doi.org/10.48550/arxiv.2310.16819} (\citeyear{https://doi.org/10.48550/arxiv.2310.16819}) focus on the estimation and statistical inference of CATE in the context of two treatments. This study assumes two linear regression models for the potential outcomes and defines CATEs as the difference between these linear regression models. The authors propose a method for consistently estimating CATEs even under high-dimensional and non-sparse parameters. They demonstrate that desirable theoretical properties, such as consistency, can be achieved without explicitly assuming sparsity. Instead, they introduce the concept of implicit sparsity, which originates from the definition of CATEs. Under this assumption, the parameters of the linear models in potential outcomes are divided into treatment-specific and common parameters. The treatment-specific parameters differ between the two models, while the common parameters remain identical. Consequently, in the difference between the two linear models, the common parameters cancel out, leaving only the differences in the treatment-specific parameters. Leveraging this assumption, the  \citeauthor{https://doi.org/10.48550/arxiv.2310.16819} (\citeyear{https://doi.org/10.48550/arxiv.2310.16819}) develop a Lasso regression method specialized for CATE estimation and show that the estimator is consistent. They further validate the soundness of their proposed method through simulation studies. This study contributes to the literature by providing a robust approach to CATE estimation in high-dimensional settings when two treatments are considered, offering practical tools for researchers dealing with complex datasets.

\subsubsection{CausalRF}

Causal Random Forests (CausalRF) are an extension of the traditional random forest algorithm tailored for estimating heterogeneous treatment effects (a.k.a. CATE), which was proposed by \citeauthor{Wager2018} (\citeyear{Wager2018}).CausalRF builds on the strengths of ensemble learning and decision trees to provide robust, nonparametric estimation of treatment effects across different subpopulations \parencite{Wager2018}.

The CausalRF algorithm involves the following key steps:
\begin{enumerate}
    \item \textbf{Building the Forest}: Grow a forest of randomized decision trees.
    \item \textbf{Splitting Criterion}: Use a splitting criterion designed to maximize the heterogeneity of treatment effects within each node.
     \item \textbf{Prediction}: Aggregate the estimates from individual trees to obtain a final CATE estimate.
\end{enumerate}

Similar to traditional random forests, CausalRF starts by constructing multiple decision trees. Each tree is grown on a bootstrap sample of the data, and at each node, a random subset of covariates is considered for splitting. The splitting criterion in CausalRF is designed to capture treatment effect heterogeneity \parencite{Wager2018}. Specifically, the criterion aims to maximize the difference in treatment effects between the resulting child nodes. One commonly used criterion is the following:
\begin{equation}
    \Delta = \left| \frac{1}{n_L} \sum_{i \in L} (Y_i - \hat{\mu}_L(\mathbf{x_i})) - \frac{1}{n_R} \sum_{i \in R} (Y_i - \hat{\mu}_R(\mathbf{x_i})) \right|,
\end{equation}
where \( L \) and \( R \) denote the left and right child nodes, \( n_L \) and \( n_R \) are the number of observations in each child node, and \( \hat{\mu}_L \) and \( \hat{\mu}_R \) are the estimated mean outcomes in each child node.

To account for treatment assignment, we can modify the criterion to focus on the difference in average treatment effects:
\begin{equation}
    \Delta = \left| \frac{1}{n_L} \sum_{i \in L} \left( \frac{Z_i Y_i}{\pi(\mathbf{x_i})} - \frac{(1 - Z_i) Y_i}{1 - \pi(\mathbf{x_i})} \right) - \frac{1}{n_R} \sum_{i \in R} \left( \frac{Z_i Y_i}{\pi(\mathbf{x_i})} - \frac{(1 - Z_i) Y_i}{1 - \pi(\mathbf{x_i})} \right) \right|,
\end{equation}
where \( Z_i \) is the treatment indicator, and \( \pi(\mathbf{x_i}) \) is the estimated propensity score.

Once the forest is built, the CATE for a new observation \( \mathbf{x_i} \) is estimated by averaging the treatment effect estimates from all trees in the forest. For each tree \( t \), the CATE estimate at \( \mathbf{x_i} \) is given by:
\begin{equation}
    \hat{\tau}_t(\mathbf{x_i}) = \frac{1}{n_t} \sum_{i \in \text{leaf}(x)} \left( \frac{Z_i Y_i}{\pi(\mathbf{x_i})} - \frac{(1 - Z_i) Y_i}{1 - \pi(\mathbf{x_i})} \right),
\end{equation}
where \( \text{leaf}(\mathbf{x_i}) \) denotes the leaf node that \( \mathbf{x_i} \) falls into, and \( n_t \) is the number of observations in that leaf.

The final CATE estimate is obtained by averaging the estimates from all trees:
\begin{equation}
    \hat{\tau}(\mathbf{x_i}) = \frac{1}{T} \sum_{t=1}^T \hat{\tau}_t(\mathbf{x_i}),
\end{equation}
where \( T \) is the total number of trees in the forest.

Recently, some important studies further developing the algorithm proposed by \citeauthor{Wager2018} (\citeyear{Wager2018}) have been performed. For instance, \citeauthor{Chen2019} (\citeyear{Chen2019}) introduces an innovative econometric procedure based primarily on the CausalRF method. This approach not only estimates the quantile treatment effect nonparametrically but also provides a measure of variable importance in terms of heterogeneity among control variables. This dual capability enhances the robustness and interpretability of causal inference models. To illustrate the practical applications of their procedure, \citeauthor{Chen2019} (\citeyear{Chen2019}) apply it to reinvestigate the distributional effect of 401(k) participation on net financial assets. They also examine the quantile earnings effect of participating in a job training program.

\citeauthor{pmlr-v97-oprescu19a} (\citeyear{pmlr-v97-oprescu19a}), on the other hand, propose the orthogonal random forest, an algorithm that combines Neyman-orthogonality to reduce sensitivity with respect to estimation error of nuisance parameters with CausalRF \parencite{Wager2018}. This method leverages the flexibility of non-parametric statistical estimation using random forests for conditional moment models. They provide a consistency rate and establish asymptotic normality for their estimator. Under mild assumptions on the consistency rate of the nuisance estimator, the authors demonstrate that their method can achieve the same error rate as an oracle with a priori knowledge of these nuisance parameters. Furthermore, \citeauthor{pmlr-v97-oprescu19a} (\citeyear{pmlr-v97-oprescu19a}) show that when nuisance functions have a locally sparse parametrization, a local $\ell_1$-penalized regression can achieve the required rate. They apply their method to estimate heterogeneous treatment effects from observational data with discrete or continuous treatments, demonstrating that their approach can control for a high-dimensional set of variables under standard sparsity conditions. The comprehensive empirical evaluation of their algorithm on both synthetic and real data showcases its robustness and practical applicability.

Finally, \citeauthor{10.1145/3437963.3441722} (\citeyear{10.1145/3437963.3441722}) addresses a critical issue in CATE estimation: the robustness to distributional shifts between training and testing data. From a causal perspective, the challenge lies in distinguishing stable causal relationships from unstable spurious correlations across different data distributions. To tackle this, they propose the Causal Transfer Random Forest (CTRF), which combines existing training data with a small amount of data from a randomized experiment to train a model resilient to feature shifts and capable of transferring to new targeting distributions. \citeauthor{10.1145/3437963.3441722} (\citeyear{10.1145/3437963.3441722}) provide theoretical justification for the robustness of the CTRF against feature shifts, leveraging knowledge from causal learning. Empirically, they evaluate the CTRF using both synthetic data experiments and real-world experiments on the Bing Ads platform, including a click prediction task and an end-to-end counterfactual optimization system. Their results show that the proposed CTRF produces robust predictions and outperforms most baseline methods in the presence of feature shifts.

\subsubsection{Neural Network Models}
\begin{figure}[H]
\centering
\setlength{\unitlength}{\textwidth}
\begin{picture}(1,0.5)
\put(-0.2,0){\includegraphics[scale=0.5]{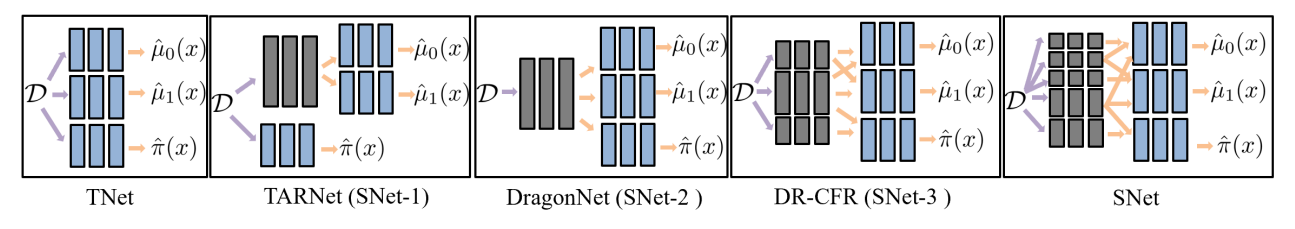}}
\end{picture}
\caption{Overview of five potential model architectures for single-step estimation of nuisance parameters, incorporating various degrees of shared information between tasks. The representation layers are depicted in gray, while the task-specific layers are illustrated in blue \parencite{pmlr-v130-curth21a}. Here $\hat{\mu_0}$ and $\hat{\mu_1}$ denote the $\hat{Y}_0$ and $\hat{Y}_1$ respectively.}
\centering
\label{Figure1}
\end{figure}

There are a great number of nonparametric models that utilize neural network architectures, especially the feed-forward neural networks (FNNs) architecture, for the estimation of CATE \parencite{pmlr-v130-curth21a}. Presumably, the most well-known models in this category are TNet \parencite{pmlr-v130-curth21a}, TARNet \parencite{pmlr-v70-shalit17a}, DragonNet \parencite{NEURIPS2019_8fb5f8be}, DR-CFR \parencite{Hassanpour2020Learning}, and SNet \parencite{pmlr-v130-curth21a}. \citeauthor{pmlr-v130-curth21a} (\citeyear{pmlr-v130-curth21a}) makes a clear representation of each architecture in their work, which can be found in this thesis in Figure \ref{Figure1}.

TNet is simply the application of T-Learner approach with FNNs for the estimation of $\hat{Y}_0$ and $\hat{Y}_1$, then CATE can be estimated using the equality $\hat{\tau}(\mathbf{x_i}) = \hat{Y}_i(1) - \hat{Y}_i(0)$. It is worth mentioning that $\pi(\mathbf{x_i})$ can also be estimated using FNNs for the TNet model. Moving to the TARNet model, it is still the application of T-Learner approach, but with the addition of a joint embedding layer (i.e., a layer that maps all data to a new/latent space) to the FNNs used for the estimation of $\hat{Y}_0$ and $\hat{Y}_1$. Yet, the estimation of $\pi(\mathbf{x_i})$ is still done with the use of the joint embedding layer. \citeauthor{NEURIPS2019_8fb5f8be} (\citeyear{NEURIPS2019_8fb5f8be}) takes the idea of using a joint embedding layer one step further, and also utilizes it for $\pi(\mathbf{x_i})$ estimation. 

DR-CFR, on the other hand, employs three distinct embedding layers, each used for modeling the propensity score, potential outcome regressions, or both. Lastly, SNet uses five embedding layers, from which three are equal to the DR-CFR model and the new two are used to allow potential-outcome regressions, $\hat{\mu_0}$ and $\hat{\mu_1}$, to
depend on only a subset of shared features. These two added embedding layers are driven by medical applications, where it is important to differentiate between markers that are prognostic (indicative of outcome regardless of treatment) and those that are predictive (indicative of treatment effectiveness) \parencite{Ballman2015}.

\citeauthor{pmlr-v130-curth21a} (\citeyear{pmlr-v130-curth21a}) demonstrated that accross different DGPs, the SNet model outperformed the other models, though when using the famous Infant Health and Development Program (IHDP) dataset \parencite{BrooksGunn1992}, the best model was DragonNet, closely followed by TARNet and DR-CFR.

\subsubsection{BART models}

Despite the neural network models being considerably powerful models for CATE estimation \parencite{pmlr-v130-curth21a}, decision-trees-based models usually outperform them for tabular data (i.e., the data usually handled in problems where the estimation of CATE and ATE is wanted) \parencite{ShwartzZiv2022,NEURIPS2022_0378c769,9998482}. Besides, the aforementioned models in this thesis only yield point-wise estimations for CATE, while usually a CATE estimation with Confidence Intervals (CI) is more informative for researchers and practitioners (as long as the CI respect the chosen level for them, say 95\%, and the CATE estimation is as accurate as the point-wise-estimations models) \parencite{CausalBook, Hahn2020}. Thus, considering both issues, the decision-tree-based model Bayesian Additive Regression Trees (BART) has come to light in the CATE estimation literature, gaining a considerable attention by researchers. Before we delve into the different CATE estimation models that use the BART model as a basis, we will first understand the architecture behind the BART model.

The BART model, introduced by \citeauthor{Chipman2010} (\citeyear{Chipman2010}), is a Bayesian approach to ensemble learning that constructs a sum-of-trees model where each tree contributes a small amount to the overall prediction. This approach allows BART to capture complex, non-linear relationships in the data.

BART represents the response variable \( Y \) as the sum of the predictions from multiple regression trees, plus an error term. Mathematically, the model can be expressed as:
\begin{equation}
    Y_i = \sum_{j=1}^m g_j(\mathbf{x_i}; T_j, M_j) + \epsilon_i, \quad \epsilon_i \sim \mathcal{N}(0, \sigma^2),
\end{equation}
where \( Y_i \) is the response variable for observation \( i \), \( g_j(\cdot) \) represents the \( j \)-th regression tree with structure \( T_j \) and terminal node parameters \( M_j \), \( \mathbf{x_i} \) are the covariates for observation \( i \), \( m \) is the total number of trees, and \( \epsilon_i \) is the normally distributed error term with mean zero and variance \( \sigma^2 \). In the Bayesian framework, priors need to be specified for the parameters of the model, including the tree structures \( T_j \), the terminal node parameters \( M_j \), and the error variance \( \sigma^2 \).

The prior for the tree structure \( T_j \) typically involves specifying probabilities for splitting rules at each node. These probabilities can be defined as:
\begin{equation}
    p(\text{split at node } t) = \alpha (1 + d_t)^{-\beta},
\end{equation}
where \( d_t \) is the depth of node \( t \), and \( \alpha \) and \( \beta \) are hyperparameters controlling the tree depth, usually set as 0.95 and 2 respectively \parencite{Chipman2010}. The prior for the terminal node parameters \( M_j \) is usually chosen to be normal:
\begin{equation}
    \mu_{j,k} \sim \mathcal{N}(\mu_{\mu}, \sigma^2_\mu),
\end{equation}
where \( \mu_{j,k} \) is the mean response at terminal node \( k \) of tree \( j \), and \( \mu_{\mu} \) and \( \sigma^2_\mu \) are hyperparameters, usually set as the solution for the problem \( m\mu_\mu - k\sqrt{m}\sigma_\mu = y_{\text{min}} \) and \( m\mu_\mu + k\sqrt{m}\sigma_\mu = y_{\text{max}} \) for a preselected value of \( k \) \parencite{Chipman2010}. For instance, selecting \( k = 2 \), which is usually the case \parencite{Chipman2010}, results in a 95\% prior probability that \( \mathbb{E}(Y | x) \) lies within the interval \( (y_{\text{min}}, y_{\text{max}}) \).

A common choice for the error variance prior is the inverse gamma distribution:
\begin{equation}
    \sigma^2 \sim \text{IG}(\nu/2, \nu \lambda/2),
\end{equation}
where \( \nu \) and \( \lambda \) are hyperparameters. To guide the specification of the hyperparameters \( \nu \) and \( \lambda \), a data-informed prior approach is usually employed. One calibrates the prior degrees of freedom \( \nu \) and scale \( \lambda \) using an approximate, data-based overestimate \( \hat{\sigma} \) of \( \sigma \). Two common choices for \( \hat{\sigma} \) include: (1) the “naive” specification, which uses the sample standard deviation of \( Y \), or (2) the “linear model” specification, which utilizes the residual standard deviation from a least squares linear regression of \( Y \) on the original \( X \) variables, with the second commonly yielding better results \parencite{Chipman2010}. A \( \nu \) value between 3 and 10 is then selected to achieve an appropriate shape and determine \( \lambda \) such that the \( q \)-th quantile of the prior on \( \sigma \) aligns with \( \hat{\sigma} \), i.e., \( P(\sigma < \hat{\sigma}) = q \). Typical \( q \) values are 0.75, 0.90, or 0.99, centering the distribution below \( \hat{\sigma} \). The default values for $(\nu,q)$ are (3,0.90) \parencite{Chipman2010}.

Lastly, the number of trees used in the BART model could also be considered a hyperparameters and is usually set to 200 \parencite{Chipman2010}.

Given the priors and the likelihood, the posterior distribution of the BART model parameters can be obtained using Markov Chain Monte Carlo (MCMC) methods. The goal is to sample from the joint posterior distribution:
\begin{equation}
    p(T_1, M_1, \ldots, T_m, M_m, \sigma^2 \mid Y, X).
\end{equation}

The MCMC algorithm for BART typically involves the following steps:
\begin{enumerate}
    \item \textbf{Initialization}: Initialize the tree structures \( T_j \), terminal node parameters \( M_j \), and error variance \( \sigma^2 \).
    \item \textbf{Gibbs Sampling}: Update each tree structure and terminal node parameters one at a time using Gibbs sampling, conditioning on the other trees and the data. Specifically, for each tree \( j \):
    \begin{itemize}
        \item Update the tree structure \( T_j \) using a Metropolis-Hastings step.
        \item Update the terminal node parameters \( M_j \) by sampling from their full conditional posterior distribution.
    \end{itemize}
    \item \textbf{Update Error Variance}: Update the error variance \( \sigma^2 \) from its full conditional posterior distribution.
    \item \textbf{Repeat}: Repeat the above steps for a large number of iterations to ensure convergence to the posterior distribution.
\end{enumerate}

Once the MCMC algorithm has converged, predictions can be made by averaging over the posterior samples of the trees.

The main CATE estimation models that use the BART model as a basis are: 1. BART model (or Vanilla BART model) \parencite{Hahn2020}, 2. BART-($f_0,f_1$) model \parencite{Hahn2020}, 3. ps-BART model \parencite{Hahn2020}, 4. BART + inverse probability of treatment weighting (IPTW) \parencite{https://doi.org/10.48550/arxiv.1905.09515}, 5. X-Learner BART \parencite{Knzel2019}.

While the Vanilla BART model is simply the BART model considering the treatment variable as "just another covariate", the ps-BART model considers both the treatment variable and estimated propensity score as "just other covariates". BART-($f_0,f_1$), on the other hand, is simply the T-Learner approach using the BART model to estimate $\hat{Y}_0$ and $\hat{Y}_1$, while X-Learner BART is the employement of the X-Learner approach using the BART model. Finally, the BART+IPTW is the Vanilla BART model with the use of the IPTW technique (for details of this technique, see \citeauthor{Chesnaye2021} (\citeyear{Chesnaye2021})).

\subsubsection{BCF}
Amidst the models using BART model as a basis, the Bayesian Causal Forest (BCF) model, proposed by \citeauthor{Hahn2020} (\citeyear{Hahn2020}), has emerged as a beacon, improving the use of the BART model for CATE estimation by proposing a novel framework that incorporates the estimated propensity score $\hat{\pi}(x_i)$ in a smart manner into the model to mitigate the negative effects of regularization-induced confounding (RIC). The essence of RIC lies in the observation that, while multiple functions may yield similar likelihood evaluations, they can imply vastly different treatment effects \parencite{Hahn2018}. This discrepancy is especially pronounced in scenarios characterized by strong confounding alongside modest treatment effects, wherein the conditional expectation of the outcome variable is predominantly influenced by covariates rather than the treatment variable \parencite{Hahn2018}, or in scenarios wherein target selection, settings where given measured covariates, treatment is assigned based on a prediction of the outcome in the absence of treatment, is present \parencite{Hahn2020}. Consequently, the posterior estimate of the treatment effect becomes significantly swayed by the prior distribution over the conditional expectation function $f(x_i, z_i)$ \parencite{Hahn2018,Hahn2020}. By explicitly including $\hat{\pi}(x_i)$, the model can account for the treatment assignment mechanism, reducing the risk of RIC. This is particularly crucial in scenarios where target selection might have occurred \parencite{Hahn2020}. The modified model incorporating the estimated propensity score is:
\begin{equation}
    \mathbb{E}[Y_i | x_i, Z_i = z_i] = \mu(x_i, \hat{\pi}(x_i)) + \tau (x_i)z_i,
\end{equation}
where $\mu(x_i, \hat{\pi}(x_i))$ represents the adjusted outcome mean, and $\mu(x_i, \hat{\pi}(x_i))$ and $\tau (x_i)$ are estimated using the BART model.

The BART priors used to estimate  $\mu(x_i, \hat{\pi}(x_i))$ and $\hat{\pi}(x_i)$ are the priors proposed by \citeauthor{Chipman2010} (\citeyear{Chipman2010}) (namely, 200 trees, $\beta$ = 2, and $\eta$ = 0.95), with one small modification made \citeauthor{Hahn2020} (\citeyear{Hahn2020}) which places a half-Cauchy prior over the scale of the leaf parameters with prior median equal to twice the marginal standard deviation of $Y$. On the other hand, the BART prior used to estimate $\tau (x_i)$ utilizes a stronger regularization, with 50 trees, $\beta$ = 3, $\eta$ = 0.25, and a half Normal prior over the scale of $\tau (x_i)$, pegging the prior median to the marginal standard deviation of $Y$ \parencite{Hahn2020}.

\citeauthor{Hahn2020} (\citeyear{Hahn2020}) demonstrated in their work that the BCF model outperforms different BART models, namely Vanilla BART model, BART-($f_0,f_1$) model, and ps-BART model, as well as CausalRF and Linear Regression regarding the metrics RMSE, Cover, and Length of CI for both CATE and ATE estimation. Nonetheless, some authors pointed out some limitations of the BCF model in the discussion part in \citeauthor{Hahn2020} (\citeyear{Hahn2020}), namely the amount of computational power that it requires and the fact that it does not scale well when adding extra covariates.

\subsection{Famous Datasets and their problems}

In the realm of causal inference, several datasets have gained prominence due to their extensive use in model-development studies and methodological evaluations \parencite{BENCHMARKS2021_2a79ea27}. Below, some of these widely recognized datasets are described:

By far the most well-known and used dataset, the IHDP dataset \parencite{BrooksGunn1992} is based on the randomized controlled experiment conducted by the Infant Health and Development Program. It comprises 747 units, with 139 treated and 608 control observations. The dataset includes 25 pre-treatment variables and a binary treatment indicator.

The Jobs dataset \parencite{pmlr-v70-shalit17a,Dehejia1999,Dehejia2002} combines data from the LaLonde experiment and the PSID comparison group. The LaLonde experimental data consists of 722 units (297 treated and 425 control), while the PSID comparison group includes 2490 control units. This dataset encompasses 8 pre-treatment variables and a binary treatment indicator.

The Twins dataset \parencite{Almond2005} is constructed from data on twin births in the USA between 1989 and 1991. It features 11,984 pairs of twins and includes 46 pre-treatment variables along with a binary treatment indicator.

The Annual Atlantic Causal Inference Conference (ACIC) Data Challenge provides a platform for comparing causal inference methodologies across various data generating processes (DGPs) \parencite{https://doi.org/10.48550/arxiv.1905.09515}. The dataset specifications vary annually. For example, ACIC 2016 included 77 datasets with 58 variables and binary treatments \parencite{Yao2021}. ACIC 2017 featured 8,000 datasets with 8 variables \parencite{Yao2021}. ACIC 2018 comprised 24 simulations with sizes ranging from 1,000 to 50,000 units and binary treatments \parencite{Yao2021}. ACIC 2019 provided 3,200 low-dimensional datasets and 3,200 high-dimensional datasets \parencite{Yao2021}.

These datasets are in theory integral to advancing the field of causal inference, offering diverse and complex scenarios for testing and validating new methodologies. Nevertheless, \citeauthor{BENCHMARKS2021_2a79ea27} (\citeyear{BENCHMARKS2021_2a79ea27}) have identified several critical flaws in the datasets commonly used for benchmarking in causal inference studies. They emphasize that the credibility of results obtained from (semi-)synthetic datasets is a significant concern. Synthetic DGPs can inherently favor specific classes of algorithms/models over others, which may not reflect realistic scenarios.

While real-world data might also favor certain algorithms, the specific biases of synthetic DGPs are often known in advance, unlike real-world scenarios where the favored algorithms are not predetermined \parencite{BENCHMARKS2021_2a79ea27}. This inherent bias limits the applicability of synthetic benchmarks, as they may not accurately simulate the conditions an algorithm will face in practical deployment. Moreover, designing effective synthetic response surfaces from scratch presents substantial challenges. The vast number of possible combinations of functional forms for outcome generation, alongside CATE-specific experimental parameters such as the degree of CATE heterogeneity and the structure of confounding, make exhaustive exploration computationally prohibitive and often impractical \parencite{BENCHMARKS2021_2a79ea27}.

Despite these limitations, \citeauthor{BENCHMARKS2021_2a79ea27} (\citeyear{BENCHMARKS2021_2a79ea27}) argue that (semi-)synthetic datasets can still be valuable for benchmarking if their inherent limitations are acknowledged and discussed appropriately when interpreting results. Simulations offer the advantage of controlled experiments, allowing researchers to directly study the effects of crucial experimental variables, such as sample size, functional forms, confounding structures, and the degree and form of effect heterogeneity.

In summary, while (semi-)synthetic datasets have inherent biases and limitations, they remain useful for understanding the performance differences of causal inference models under various controlled conditions. The key lies in being transparent about these limitations and carefully considering them during result interpretation.

\subsection{Application of Causal Inference in research and industry}
There are a great number of empirical studies that leverage CATE estimation methods to create knowledge about causal effects of certain variables on complex problems. One example is the work of \citeauthor{https://doi.org/10.48550/arxiv.1812.04345} (\citeyear{https://doi.org/10.48550/arxiv.1812.04345}), which focuses on understanding gender inequality in earnings in the United States.

In their study, \citeauthor{https://doi.org/10.48550/arxiv.1812.04345} (\citeyear{https://doi.org/10.48550/arxiv.1812.04345}) analyze data from the 2016 American Community Survey using a high-dimensional wage regression and applying double LASSO to quantify heterogeneity in the gender wage gap. They found that, in 2016, the majority of full-time employed women in the U.S. earned significantly less than comparable men. However, the extent of this wage gap varied greatly depending on socio-economic characteristics such as marital status, educational attainment, race, occupation, and industry. The study revealed that the gender wage gap was driven primarily by factors like marital status, having children at home, race, occupation, industry, and educational attainment. These findings highlight the complex and multifaceted nature of gender inequality in earnings, suggesting that simple, one-size-fits-all policies may be insufficient to address the issue effectively. 

The insights provided by this research are crucial for policymakers, as they emphasize the importance of considering various socio-economic factors when designing policies aimed at reducing discrimination and unequal pay. This study exemplifies how causal inference methods, particularly CATE estimation, can be utilized to uncover important socio-economic dynamics and inform policy decisions.

Another example is the study of \citeauthor{Yeager2019} (\citeyear{Yeager2019}), which assesses the impact of behavioral interventions on academic outcomes. \citeauthor{Yeager2019} (\citeyear{Yeager2019}) utilizes the BCF model to conduct a comprehensive Bayesian analysis of the effectiveness of a brief, online growth mindset intervention designed to teach students that intellectual abilities can be developed. This intervention was tested on a nationally representative sample of secondary school students in the United States. The study's findings revealed that the growth mindset intervention had a significant positive impact on the academic performance of lower-achieving students, particularly in improving their grades. Additionally, the intervention led to an increase in overall enrollment in advanced mathematics courses. A key insight from this research was the identification of school contexts that amplified the intervention's effects, specifically where peer norms were aligned with the intervention's message.

The robustness of \citeauthor{Yeager2019}'s (\citeyear{Yeager2019}) conclusions was bolstered by several methodological strengths, including independent data collection and processing, pre-registration of analyses, and the use of a blinded Bayesian analysis. The application of the BCF model allowed for a nuanced understanding of the intervention's effectiveness across different contexts thanks to the CATE estimation of the intervention. 

Another study that beneficiated from the CATE estimation with the help of the BCF model is the work of \citeauthor{Yeager2022} (\citeyear{Yeager2022}). \citeauthor{Yeager2022} (\citeyear{Yeager2022}) have evaluated the effectiveness a novel intervention aimed at improving adolescents' responses to social-evaluative stressors. These stressors, which involve the fear of negative judgment, pose a significant threat to adolescent mental health and can lead to disengagement from stressful but beneficial activities. The study introduced a short, approximately 30-minute, scalable intervention called the "synergistic mindsets" intervention. This self-administered online training module targets both growth mindsets (the belief that intelligence can be developed) and stress-can-be-enhancing mindsets (the belief that physiological stress responses can enhance performance). Across six double-blind, randomized, controlled experiments involving secondary and post-secondary students in the United States, the intervention showed replicable benefits.

Key findings from the study included improvements in stress-related cognitions, cardiovascular reactivity, daily cortisol levels, psychological well-being, academic success, and anxiety symptoms during the 2020 COVID-19 lockdowns. Heterogeneity analyses and a four-cell experiment demonstrated that the intervention's benefits were maximized when both growth and stress mindsets were addressed synergistically.

Finally, \citeauthor{Bail2019} (\citeyear{Bail2019}) have studied the impact of social-media influence campaigns on public opinion and political behavior through ATE and CATE estimation, anew using the BCF model. They have investigated the effects of interactions with Twitter accounts operated by the Russian Internet Research Agency (IRA) on American political attitudes and behaviors. The study combined longitudinal data from 1,239 Republican and Democratic Twitter users collected in late 2017 with nonpublic data from Twitter on IRA-operated accounts. Despite widespread concerns about the influence of Russian social-media campaigns on political divisions in the United States, \citeauthor{Bail2019} (\citeyear{Bail2019}) found no substantial evidence that interactions with IRA accounts significantly impacted six key measures of political attitudes and behaviors over a one-month period. The study concluded that while the IRA's social-media campaign might not have effectively altered political attitudes or behaviors in the short term, important questions remain regarding the broader implications of such campaigns on misinformation, political discourse, and the dynamics of the 2016 presidential election. \citeauthor{Bail2019} (\citeyear{Bail2019}) emphasized the need for further research to fully understand the influence of social-media campaigns on political polarization and to address the limitations of their study, including the inability to determine the impact of IRA accounts on the 2016 election.

\section{K-Fold Causal BART}\label{K-Fold Causal BART}

The proposed model, K-Fold Causal BART, can be divided into two parts: the model for ATE estimation and the model for CATE estimation. Below, a detailed explanation for both parts of the proposed model can be found.

\subsection{ATE Estimation}

For the ATE estimation, the proposed model employs the Double/Orthogonal machine learning method for the Partially Linear Model \parencite{CausalBook}. This approach is can be described as:

\begin{enumerate}
    \item \textbf{Data Partitioning}:
    \begin{itemize}
        \item Randomly divide the data indices into \( K \) folds of approximately equal size, such that \( \{1, \ldots, n\} = \bigcup_{k=1}^K I_k \). Heuristically, five folds as a default is usually an optimal choice for most datasets \parencite{CausalBook}.
        \item For each fold \( k \) (where \( k = 1, \ldots, K \)), estimate the functions \(\hat{\ell}^{[k]}\) and \(\hat{m}^{[k]}\) for the conditional expectation functions \(\ell\) and \(m\), excluding the \(k\)-th fold of data, using the standard BART model (i.e., the BART model with its default hyperparameters and priors) as a basis for the regression.
        \item Calculate the cross-fitted residuals for each \( i \in I_k \) as:
        \[
        \dot{Y}_i = Y_i - \hat{\ell}^{[k]}(X_i), \quad \dot{D}_i = D_i - \hat{m}^{[k]}(X_i).
        \]
    \end{itemize}
    
    \item \textbf{Ordinary Least Squares (OLS) Regression}:
    \begin{itemize}
        \item Perform an OLS regression of \(\dot{Y}_i\) on \(\dot{D}_i\).
        \item Determine \(\hat{\beta}\), which would be the ATE estimation, as the solution to the normal equations:
        \[
        \mathbb{E}_n \left[ (\dot{Y} - b \dot{D}) \dot{D} \right] = 0.
        \]
    \end{itemize}
    
    \item \textbf{Inference}:
    \begin{itemize}
        \item Construct standard errors and Confidence Intervals (CIs) following the principles of standard least squares theory. Hence, the construct CIs will be frequentist, and not Bayesian in this case. Nonetheless, a Bayesian approach would also be possible for the linear regression part of the ATE estimation.
    \end{itemize}
\end{enumerate}

\subsection{CATE Estimation}

To estimate CATE, a K-Fold T-Learner approach is used for the K-Fold Causal BART model, following the steps below:

\begin{enumerate}
    \item \textbf{Data Partitioning}:
    \begin{itemize}
        \item Divide the data into $K$ folds. This allows for multiple iterations where each fold is used as a validation set while the remaining $K-1$ folds are used for training. Heuristically, five folds as a default is usually an optimal choice for most datasets \parencite{CausalBook}.
    \end{itemize}

    \item \textbf{Cross-Fitted Predictions}:
    \begin{itemize}
        \item For each individual in the validation fold $k$, use the standard BART models (i.e., the BART model with its default hyperparameters and priors) trained on the other $K-1$ folds to predict the outcomes, reducing overfitting and providing more robust estimates:
        \[
        \hat{Y}_1(X) = \hat{Y}_1^{[k]}(X) \quad \text{and} \quad \hat{Y}_0(X) = \hat{Y}_0^{[k]}(X)
        \]
        It is worth mentioning that posterior simulations for $\hat{Y}_1(X)$ and $\hat{Y}_0(X)$ are still not used in this stage, but rather only their mean prediction to accelerate the entire K-Fold Causal BART model training.
    \end{itemize}
    
    \item \textbf{Estimating CATE}:
    \begin{itemize}
        \item Compute the CATE for each individual as:
        \[
        \dot{\tau}(X) = \hat{Y}_1(X) - \hat{Y}_0(X)
        \]
        \item To further debias the CATE estimates $\dot{\tau}(X)$ obtained from the T-Learner approach, another K-Fold regression of $\dot{\tau}(X)$ on $X$ is performed using the standard BART model (i.e., the BART model with its default hyperparameters and priors):
        \[
        \dot{\tau}(X) = \hat{\tau}(X) + \epsilon
        \]
        Here, on the other hand, posterior simulations for $\hat{\tau}(X)$ are used to construct CIs. Nonetheless, given the the mean predictions of $\hat{Y}_1(X)$ and $\hat{Y}_0(X)$ were used, instead of all simulations using the MCMC method for example, the CIs of $\hat{\tau}(X)$ are systematically overconfident. As a result, a heuristical solution for this issue, namely multiplying the CI length by 1.5, has been devised and used. Such problem-solving solution can be compared to the use of tempered posteriors in Bayesian machine learning contexts \parencite{pmlr-v222-pitas24a, NEURIPS2022_73e018a0, https://doi.org/10.48550/arxiv.2008.00029}.
    \end{itemize}
\end{enumerate}

The dual application of K-Fold cross-validation enhances the robustness and accuracy of the final CATE estimates, leading to improved performance and more reliable insights. This is the case as by regressing $\dot{\tau}(X)$ on $X$, we are effectively fitting a meta-model to capture and get ridge off (in any case partially) any systematic biases as the initial estimates $\dot{\tau}(X)$ might still contain some prediction errors or biases due to model misspecification, heterogeneity in the data, or limited sample sizes.

\section{Research Design}\label{Research Design}

\subsection{Data Generating Processes and Datasets}

The empirical validation of the proposed model for CATE estimation is conducted using both synthetic and semi-synthetic datasets. Synthetic datasets will be generated based on specified Data Generating Processes (DGPs), while semi-synthetic datasets, such as the IHDP benchmark, will provide a real-world context for model evaluation.

\subsubsection*{Synthetic DGPs}

The synthetic DGPs are designed to simulate various scenarios of treatment effects and covariate distributions. The following equations define the DGPs used in this study, which are equal to the DGPs used by \citeauthor{Hahn2020} (\citeyear{Hahn2020}):

\begin{itemize}
    \item \textbf{Homogeneous Linear:}
    \begin{equation}
        \tau(x) = 3, \quad \mu(x) = 1 + g(x_4) + x_1 x_3,
    \end{equation}
    where $g(1) = 2, g(2) = -1, g(3) = -4$, $x_1$ and $x_3 \sim \mathcal{N}(0,1)$, and $x_4$ is unordered categorical, taking three levels (denoted 1, 2, 3).
    
    \item \textbf{Heterogeneous Linear:}
    \begin{equation}
        \tau(x) = 1 + 2x_2 x_5,
    \end{equation}
    with $\mu(x)$ as in the homogeneous case, $x_2 \sim \mathcal{N}(0,1)$ and $x_5$ is a dichotomous variable.
    
    \item \textbf{Homogeneous Nonlinear:}
    \begin{equation}
        \mu(x) = -6 + g(x_4) + 6 |x_3 - 1|,
    \end{equation}
    with $\tau(x)$ as in the homogeneous linear case.
    
    \item \textbf{Heterogeneous Nonlinear:} $\mu(x)$ as in the homogeneous nonlinear case, with $\tau(x)$ as in the heterogeneous linear case.
    
    \item \textbf{Propensity Function:}
    \begin{equation}
        \pi(x_i) = 0.8 \Phi \left( \frac{3\mu(x_i)}{s} - 0.5x_1 \right) + 0.05 + \frac{u_i}{10},
    \end{equation}
    where $s$ is the standard deviation of $\mu(x)$ over the observed sample, and $u_i \sim \text{Uniform}(0, 1)$.
\end{itemize}

Each DGP will be simulated with different sample sizes ($n = 250$ and $500$) to assess the scalability and robustness of the models. Each scenario will be replicated 100 times, following the procedure outlined in \citeauthor{Hahn2020} (\citeyear{Hahn2020}), to ensure statistical reliability and robustness in the findings.

\subsubsection*{Semi-Synthetic Dataset}

The IHDP benchmark dataset \parencite{BrooksGunn1992}, a widely recognized semi-synthetic dataset in causal inference research, will be utilized to validate the models in a 'real-world' context. Originating from the Infant Health and Development Program, this dataset provides a comprehensive set of real-world covariates ($n = 747$, $d = 25$) derived from a randomized experiment designed to assess the impact of specialized childcare interventions on premature infants with low birth weight. The initial findings, reported by \citeauthor{BrooksGunn1992} (\citeyear{BrooksGunn1992}), underscored a significant positive effect on cognitive test scores, showcasing the potential benefits of targeted early-life interventions.

To adapt this dataset for causal inference research, particularly to study observational study dynamics, \citeauthor{Hill2011} (\citeyear{Hill2011}) introduced an element of confounding and imbalance by selectively excluding a subset of treated individuals—specifically, those whose mothers were nonwhite. This manipulation resulted in an uneven distribution between the treatment ($n_1 = 139$) and control groups ($n_0 = 608$), effectively simulating the incomplete overlap often encountered in observational studies and introducing challenges akin to those faced in real-world data scenarios.

The IHDP benchmark uses a specific DGP, referred to as setup 'B' by \citeauthor{Hill2011} (\citeyear{Hill2011}), which is defined by the equations:
\begin{equation}
\mu_0(x) = \exp((x + A)\beta), \quad \mu_1(x) = x\beta - \omega,
\end{equation}
where $\beta$ represents a sparse coefficient vector with values drawn from the set $\{0, 0.1, 0.2, 0.3, 0.4\}$ with respective probabilities $\{0.6, 0.1, 0.1, 0.1, 0.1\}$, $A$ denotes a fixed offset matrix, and $\omega$ is adjusted in each simulation to ensure that the Average Treatment Effect on the Treated (ATT) remains consistent at 4. This careful calibration allows for the controlled study of treatment effects while accounting for the intricacies of confounding and selection bias inherent in observational data.

Further expanding on \citeauthor{Hill2011} (\citeyear{Hill2011})'s work, \citeauthor{pmlr-v70-shalit17a} (\citeyear{pmlr-v70-shalit17a}) enhanced the benchmark by generating 100 realizations of the DGP. This augmentation not only facilitates the robust evaluation of causal inference models across multiple simulated scenarios but also mirrors the variability and unpredictability encountered in empirical data, thereby providing a rigorous testing ground for methodologies. As a result, the benchmark dataset as proposed by \citeauthor{pmlr-v70-shalit17a} (\citeyear{pmlr-v70-shalit17a}) is used in this research.

\subsection{Model Evaluation and Validation}

This research project employs a multifaceted approach to model evaluation, incorporating both error measures and uncertainty metrics, inspired by \citeauthor{Hahn2020} (\citeyear{Hahn2020}). The error measure chosen for the evaluation of CATE and ATE estimation is the Root Mean Square Error (RMSE), defined as:
    \begin{equation}
        RMSE = \sqrt{\frac{1}{n} \sum_{i=1}^{n} (\hat{y}_i - y_i)^2},
    \end{equation}
where $\hat{y}_i$ is the mean predicted CATE or ATE value, $y_i$ is the actual CATE or ATE value, and $n$ is the sample size. To evaluate the model's uncertainty estimation, the following model uncertainty performance metrics are utilized for both CATE and ATE:

\begin{itemize}
    \item \textbf{Coverage:} The proportion of times the true value falls within the predicted credible interval. Ideal coverage for a 95\% credible interval should be close to 95\%.
    
    \item \textbf{Interval Length:} The average width of the credible intervals, serving as a critical indicator of the precision and reliability of uncertainty estimates in Bayesian models. A well-calibrated Bayesian model should produce credible intervals that are sufficiently narrow to be informative yet wide enough to encompass the true parameter values, reflecting an accurate representation of the underlying uncertainty. This balance is crucial in practical applications where decision-making relies on the precision of predictive models.
\end{itemize}

\subsection{Benchmark Models}
The benchmark models chosen for this research are inspired by \citeauthor{Hahn2020} (\citeyear{Hahn2020})'s benchmark models selections are composed of: BCF, ps-BART, BART-$(f_0, f_1)$, Causal RF, and regularized linear model (LM), using the horseshoe prior (HR) of \citeauthor{Carvalho2010} (\citeyear{Carvalho2010}), with up to three way interactions.

\subsection{Ablation Study}
To better understand where the advantages and disadvantages of the K-Fold Causal BART model come from, an ablation study is performed for both the synthetic datasets and the semi-synthetic dataset. The sub-models of the K-Fold Causal BART model and their respective aim for the ablation study is described below:

\begin{enumerate}
    \item Without $\mathbb{E}[\dot{\tau}(X)]$: To study the extent to which the employment of another K-Fold regression of $\dot{\tau}(X)$ on $X$ to further debias and improve the CATE estimation of the proposed model actually further debiases and improves the CATE estimation.
    \item Without Partialling Out and LM: To study the extent to which the use of the Double/Orthogonal machine learning method for the Partially Linear Model for ATE estimation actually improves the accuracy of the ATE estimation.
    \item Without K-Fold for $\mathbb{E}[\dot{\tau}(X)]$: To study the extent to which the utilization of the K-Fold method for the regression of $\dot{\tau}(X)$ on $X$ actually debiases and improves the precision of the CATE estimation.
    \item Without K-Fold for $\hat{Y}_1(X)$ and $\hat{Y}_0(X)$: To study the extent to which the employment of the K-Fold method for the T-Learner part of the K-Fold Causal BART model actually improves the accuracy of the CATE estimation.
\end{enumerate}

\section{Results}\label{Results}
As already stated in the previous section, in this section the simulations results for the synthetic datasets as well as for the famous IHDP dataset are discussed. Additionally, for each dataset, an ablation study is performed and its results analysed to further understand which components of the K-Fold Causal BART model are responsible for its performance for a certain metric.

\subsection{Synthetic Datasets}

Table \ref{table1} presents the results for the linear synthetic DGPs. As already commented by \citeauthor{Hahn2020} (\citeyear{Hahn2020}), the LM + HS model is the most powerful model, which is due to the fact that the synthetic DGPs here favor the LM + HS model due to their data generating nature rather than the fact that the LM + HS model is better than the other models \parencite{BENCHMARKS2021_2a79ea27}. When compared to the other models, the K-Fold Causal BART shows itself to be a competitive model when considering the homogeneous effect of the treatment, though the cover for both CATE and especially ATE is inferior to the BART models. Moving to the heterogeneous treatment effect DGP, then the K-Fold Causal BART becomes more powerful than the other models for both ATE and CATE estimation, though the cover for ATE is still systematically far below 0.95. The results regarding the cover for ATE could potentially show that the choice of the frequentist CIs instead of the Bayesian CIs could have been a bad choice.

In Table \ref{table2}, the results for the nonlinear synthetic DGPs can be found. Here, it is clear that the LM + HS model extraordinary performance for the other synthetic DGPs was indeed due to the fact that they favor the LM + HS model. Anew the cover for ATE is systematically below 0.95 for the K-Fold Causal BART model, showing an overconfidence of the frequentist CIs. Nonetheless, this time the ATE estimation for the homogeneous treatment effect DGP of the K-Fold Causal BART model is significantly superior to the other models, while the CATE estimation remaining competitive to the BCF model which is the top model regarding this metric for the homogeneous treatment effect DGP. It is worth mentioning that even though the CATE estimation of the K-Fold Causal BART model is comparable to the BCF model, the BCF model clearly gives better CIs (i.e., with cover closer to 0.95 while having a smaller length). Now concerning the heterogeneous treatment effect DGP, both the ATE and CATE are superior to all other models considering RMSE (and cover and length for CATE). Consequently, the results of the synthetic datasets demonstrate the competitiveness of the proposed model in this research, with it being as accurate as the other top-performing models in most cases and in some cases even more accurate. However, the use of frequentist CIs for the ATE estimation of the proposed model showed itself to presumably not be the best choice, though the use of the ablation study could confirm or falsify this hypothesis.

\begin{landscape}

\begin{table}[ht]
\centering
\caption{Simulation study results when the true DGP is a linear model.}
\begin{tabular}{llcccccccccccc}
\hline
& & \multicolumn{5}{c}{Homogeneous effect} & \multicolumn{7}{c}{Heterogeneous effects} \\
\cline{3-14}
$n$ & Method & \multicolumn{2}{c}{ATE} & & \multicolumn{2}{c}{CATE} & \multicolumn{4}{c}{ATE} & & \multicolumn{1}{c}{CATE} \\
\cline{3-5} \cline{6-8} \cline{9-10} \cline{11-14}
& & RMSE & cover & length & RMSE & cover & length & RMSE & cover & length & RMSE & cover & length \\
\hline
250 & K-Fold Causal BART & 0.25& 0.31 & 0.19 & 0.49 & 0.99 & 4.0 & 0.21 & 0.69 & 0.27 & 0.70 & 0.95 & 2.73 \\
& BCF & 0.21 & 0.92 & 0.91 & 0.48 & 0.96 & 2.0 & 0.27 & 0.84 & 0.99 & 1.09 & 0.91 & 3.3 \\
& ps-BART & 0.22 & 0.94 & 0.97 & 0.44 & 0.99 & 2.3 & 0.31 & 0.90 & 1.13 & 1.30 & 0.89 & 3.5 \\
& BART $(f_0, f_1)$ & 0.56 & 0.41 & 0.99 & 0.92 & 0.93 & 3.4 & 0.61 & 0.44 & 1.14 & 1.47 & 0.90& 4.5 \\
& Causal RF & 0.34 & 0.73 & 0.98 & 0.47 & 0.84 & 1.3 & 0.49 & 0.68 & 1.25 & 1.58 & 0.68& 2.4 \\
& LM + HS & 0.14 & 0.96 & 0.83 & 0.26 & 0.99 & 1.7 & 0.17 & 0.94 & 0.89 & 0.33 & 0.99 & 1.9 \\
\hline
500 & K-Fold Causal BART & 0.10 & 0.56 & 0.11 & 0.31 & 0.99 & 3.65 & 0.11 & 0.78 & 0.17 & 0.44 & 0.98 & 2.34 \\
& BCF & 0.16 & 0.88 & 0.60 & 0.38 & 0.95 & 1.4 & 0.16 & 0.90 & 0.64 & 0.79 & 0.89& 2.4 \\
& ps-BART & 0.18 & 0.86 & 0.63 & 0.35 & 0.99 & 1.8 & 0.16 & 0.90 & 0.69 & 0.86 & 0.95& 2.8 \\
& BART-$(f_0, f_1)$ & 0.47 & 0.21 & 0.66 & 0.80 & 0.93 & 3.1 & 0.42 & 0.42 & 0.75 & 1.16 & 0.92& 3.9 \\
& Causal RF & 0.36 & 0.47 & 0.69 & 0.52 & 0.75 & 1.2 & 0.40 & 0.59 & 0.88 & 1.30 & 0.71& 2.1 \\
& LM + HS & 0.11 & 0.96 & 0.54 & 0.18 & 0.99 & 0.12 & 0.93 & 0.59 & 0.22 & 0.22 & 0.98& 1.2 \\
\hline
\end{tabular}
\label{table1}
\end{table}
    
\begin{table}[ht]
\centering
\caption{ Simulation study results when the true DGP is nonlinear.}
\begin{tabular}{llcccccccccccc}
\hline
& & \multicolumn{5}{c}{Homogeneous effect} & \multicolumn{7}{c}{Heterogeneous effects} \\
\cline{3-14}
$n$ & Method & \multicolumn{2}{c}{ATE} & & \multicolumn{2}{c}{CATE} & \multicolumn{4}{c}{ATE} & & \multicolumn{1}{c}{CATE} \\
\cline{3-5} \cline{6-8} \cline{9-10} \cline{11-14}
& & RMSE & cover & length & RMSE & cover & length & RMSE & cover & length & RMSE & cover & length \\
\hline
250 & K-Fold Causal BART & 0.11 & 0.86 & 0.21 & 0.79 & 0.99 & 4.36 & 0.14 & 0.885 & 0.29 & 0.94 & 0.93 & 2.98 \\
& BCF & 0.26 & 0.945 & 1.3 & 0.63 & 0.94 & 2.5 & 0.30 & 0.930 & 1.4 & 1.3&0.93& 4.5 \\
& ps-BART & 0.54 & 0.780 & 1.6 & 1.00 & 0.96 & 4.3 & 0.56 & 0.805 & 1.7 & 1.7 & 0.91 & 5.4 \\
& BART $(f_0, f_1)$ & 1.48 & 0.035 & 1.5 & 2.42 & 0.80 & 6.4 & 1.44 & 0.085 & 1.6 & 2.6 & 0.83 & 7.1 \\
& Causal RF & 0.81 & 0.425 & 1.5 & 0.84 & 0.70 & 2.0 & 1.10 & 0.305 & 1.8 &  1.8 & 0.66 & 3.4 \\
& LM + HS & 1.77 & 0.015 & 1.8 & 2.13 & 0.54 & 4.4 & 1.65 & 0.085 & 1.9 & 2.2 & 0.62 & 4.8 \\
\hline
500 & K-Fold Causal BART & 0.07 & 0.83 & 0.12 & 0.49 & 0.99 & 3.74 & 0.09 & 0.885 & 0.19 & 0.54 & 0.97 & 2.38 \\
& BCF & 0.20 & 0.945 & 0.97 & 0.47 & 0.94 & 1.9 & 0.23 & 0.910 & 0.97 & 1.0 & 0.92 & 3.4 \\
& ps-BART & 0.24 & 0.910 & 1.07 & 0.62 & 0.99 & 3.3 & 0.26 & 0.890 & 1.06 & 1.1 & 0.95 & 4.1 \\
& BART-$(f_0, f_1)$ & 1.11 & 0.035 & 1.18 & 2.11 & 0.81 & 5.8 & 1.09 & 0.065 & 1.17 & 2.3 & 0.82 & 6.2 \\
& Causal RF & 0.39 & 0.650 & 1.00 & 0.54 & 0.87 & 1.7 & 0.59 & 0.515 & 1.18 & 1.5 & 0.73 & 2.8 \\
& LM + HS & 1.76 & 0.005 & 1.34 & 2.19 & 0.40 & 3.5 & 1.71 & 0.000 & 1.34 & 2.2 & 0.45 & 3.7 \\
\hline
\end{tabular}
\label{table2}
\end{table}

\end{landscape}

Table \ref{table3} shows the ablation study results for the linear DGP. Starting with the first sub-model, Without $\mathbb{E}[\dot{\tau}(X)]$, it can be seen that the utilization of another K-Fold regression of $\dot{\tau}(X)$ on $X$ indeed improves the CATE estimation considering all metrics. Moving to the second sub-model, Without Partialling Out and LM, the employment of the Double/Orthogonal machine learning method for the Partially Linear Model for ATE estimation slightly improves the accuracy of ATE estimation while decreasing the performance in properly creating CIs for it. Regarding the third sub-model, Without K-Fold for $\mathbb{E}[\dot{\tau}(X)]$, the use of the K-fold method for the regression of $\dot{\tau}(X)$ on $X$ does not seem to improve the model performance at all, while adding computational costs to it. Hence, if only considering the results of the linear DGP, one would conclude that the employment of the K-fold method for the regression of $\dot{\tau}(X)$ on $X$ should be removed from the proposed model. Finally, concerning the last sub-model, Without K-Fold for $\hat{Y}_1(X)$ and $\hat{Y}_0(X)$, it can be observed that the utilization of the K-Fold method for the T-Learner part greatly improves the ATE estimation accuracy while seemingly having no impact on the precision of the CATE estimation. Consequently, we could affirm that since the CATE estimation has a second use of the K-Fold method, the lack of the first one does not negatively impact its estimation. In other words, it could be affirmed that a second K-fold method is not necessary to further avoid overfitting of the model, which would also explain the results for the third sub-model.

The results for the nonlinear DGP can be seen in Table \ref{table4}. Similarly to the results of the linear DGP, the results of Table \ref{table4} indicate that the employment of another K-Fold regression of $\dot{\tau}(X)$ on $X$ indeed improves the CATE estimation concerning all performance metrics. Moving to the second sub-model, it can be observed that the use of the Double/Orthogonal machine learning method for the Partially Linear Model for ATE estimation considerably improves the accuracy of ATE estimation while not having a significant impact on the performance of CIs creation. Concerning the third sub-model, the employment of the K-fold method for the regression of $\dot{\tau}(X)$ on $X$ anew does not seem to improve the model performance at all, while adding computational costs to it. Also similarly to the results of the linear DGP, the use of the K-Fold method for the T-Learner part greatly improves the ATE estimation accuracy while having a slightly negative impact on the precision of the CATE estimation. Again, the hypothesis that a second K-fold method is not necessary to further avoid overfitting of the model can be raised.

\begin{landscape}

\begin{table}[ht]
\caption{Ablation study results when the true DGP is a linear model}
\centering
\begin{tabular}{llcccccccccccc}
\hline
& & \multicolumn{5}{c}{Homogeneous effect} & \multicolumn{7}{c}{Heterogeneous effects} \\
\cline{3-14}
$n$ & Method & \multicolumn{2}{c}{ATE} & & \multicolumn{2}{c}{CATE} & \multicolumn{4}{c}{ATE} & & \multicolumn{1}{c}{CATE} \\
\cline{3-5} \cline{6-8} \cline{9-10} \cline{11-14}
& & RMSE & cover & length & RMSE & cover & length & RMSE & cover & length & RMSE & cover & length \\
\hline
250 & K-Fold Causal BART & 0.25& 0.31 & 0.19 & 0.49 & 0.99 & 4.00 & 0.21 & 0.69 & 0.27 & 0.70 & 0.95 & 2.73 \\
& Without $\mathbb{E}[\dot{\tau}(X)]$ & 0.25 & 0.31 & 0.19 & 0.64 & 0.99 & 5.69 & 0.21 & 0.69 & 0.27 & 0.77 & 0.99 & 4.79 \\
& Without Partialling Out and LM & 0.26 & 0.00 & 0.18 & 0.49 & 0.99 & 4.00 & 0.27 & 0.99 & 0.26 & 0.70 & 0.95 & 2.73 \\
& Without K-Fold for $\mathbb{E}[\dot{\tau}(X)]$ & 0.25& 0.31 & 0.19 & 0.50 & 0.99 & 3.88 & 0.21 & 0.69 & 0.27 & 0.69 & 0.92 & 2.38 \\
& Without K-Fold for $\hat{Y}_1(X)$ and $\hat{Y}_0(X)$ & 0.57 & 0.01 & 0.11 & 0.49 & 0.99 & 3.87 & 0.44 & 0.01 & 0.11 & 0.60 & 0.96 & 2.54 \\

\hline
500 & K-Fold Causal BART & 0.10 & 0.56 & 0.11 & 0.31 & 0.99 & 3.65 & 0.11 & 0.78 & 0.17 & 0.44 & 0.98 & 2.34 \\
& Without $\mathbb{E}[\dot{\tau}(X)]$ & 0.10 & 0.56 & 0.11 & 0.46 & 0.99 & 4.50 & 0.11 & 0.78 & 0.17 & 0.53 & 0.78 & 3.30 \\
& Without Partialling Out and LM & 0.11 & 1.00 & 0.11 & 0.31 & 0.99 & 3.65 & 0.12 & 0.98 & 0.16 & 0.44 & 0.98 & 2.34 \\
& Without K-Fold for $\mathbb{E}[\dot{\tau}(X)]$ & 0.10 & 0.56 & 0.11 & 0.32 & 0.99 & 3.58 & 0.11 & 0.78 & 0.17  & 0.44 & 0.97 & 2.10 \\
& Without K-Fold for $\hat{Y}_1(X)$ and $\hat{Y}_0(X)$ & 0.53 & 0.00 & 0.07 & 0.32 & 0.99 & 3.54 & 0.36 & 0.00 & 0.08 & 0.39 & 0.98 & 2.13 \\
\hline
\end{tabular}
\label{table3}
\end{table}

\begin{table}[ht]
\centering
\caption{Ablation study results when the true DGP is nonlinear.}
\begin{tabular}{llcccccccccccc}
\hline
& & \multicolumn{5}{c}{Homogeneous effect} & \multicolumn{7}{c}{Heterogeneous effects} \\
\cline{3-14}
$n$ & Method & \multicolumn{2}{c}{ATE} & & \multicolumn{2}{c}{CATE} & \multicolumn{4}{c}{ATE} & & \multicolumn{1}{c}{CATE} \\
\cline{3-5} \cline{6-8} \cline{9-10} \cline{11-14}
& & RMSE & cover & length & RMSE & cover & length & RMSE & cover & length & RMSE & cover & length \\
\hline
250 & K-Fold Causal BART & 0.11 & 0.86 & 0.21 & 0.79 & 0.99 & 4.36 & 0.14 & 0.885 & 0.29 & 0.94 & 0.93 & 2.98 \\
& Without $\mathbb{E}[\dot{\tau}(X)]$ & 0.11 & 0.86 & 0.21 & 0.95 & 0.99 & 7.49 & 0.14 & 0.885 & 0.29 & 1.06 & 0.98 & 7.02 \\
& Without Partialling Out and LM & 0.34 & 1.00 & 0.22 & 0.79 & 0.99 & 4.36 & 0.38 & 0.88 & 0.29 & 0.94 & 0.93 & 2.98 \\
& Without K-Fold for $\mathbb{E}[\dot{\tau}(X)]$ & 0.11 & 0.86 & 0.21 & 0.83 & 0.98 & 4.13 & 0.14 & 0.885 & 0.29 & 0.97 & 0.90 & 2.56 \\
& Without K-Fold for $\hat{Y}_1(X)$ and $\hat{Y}_0(X)$ & 1.10 & 0.00 & 0.11 & 0.65 & 0.99 & 3.90 & 0.55 & 0.00 & 0.10 & 0.78 & 0.95 & 2.58 \\
\hline
500 & K-Fold Causal BART & 0.07 & 0.83 & 0.12 & 0.49 & 0.99 & 3.74 & 0.09 & 0.88 & 0.19 & 0.54 & 0.97 & 2.38 \\
& Without $\mathbb{E}[\dot{\tau}(X)]$ & 0.07 & 0.83 & 0.12 & 0.62 & 0.99 & 5.16 & 0.09 & 0.88 & 0.19 & 0.66 & 0.99 & 4.12 \\
& Without Partialling Out and LM & 0.13 & 1.00 & 0.11 & 0.49 & 0.99 & 3.74 & 0.13 & 0.98 & 0.18 & 0.54 & 0.97 & 2.38 \\
& Without K-Fold for $\mathbb{E}[\dot{\tau}(X)]$ & 0.07 & 0.83 & 0.12 & 0.52 & 0.99 & 3.61 & 0.09 & 0.88 & 0.19 & 0.56 & 0.97 & 2.38 \\
& Without K-Fold for $\hat{Y}_1(X)$ and $\hat{Y}_0(X)$ & 0.71 & 0.00 & 0.08 & 0.43 & 0.99 & 3.52 & 0.38 & 0.01 & 0.09 & 0.47 & 0.97 & 2.09 \\
\hline
\end{tabular}
\label{table4}
\end{table}

\end{landscape}

\subsection{Semi-Synthetic Datasets}
\subsubsection{Aggregated Results Analysis}

Table \ref{table5} presents the results of the experiments related to the IHDP dataset. Considering ATE, it can be observed that the proposed model is interestingly the worst model, while the BART-$(f_0, f_1)$ model and ps-BART model are the best models. However, given the competitive results of ATE estimation considering RMSE for the synthetic datasets, it could be hypothesized that the IHDP dataset simply does not favor the use of the Double/Orthogonal machine learning method for the Partially Linear Model for ATE estimation. Moving to CATE estimation, the BART-$(f_0, f_1)$ model and ps-BART model remain the best models while the proposed model has a similar performance to the BCF model, though the CIs of the BCF model are clearly superior given their considerably smaller length. Interestingly, the results contradict the results of \citeauthor{Hahn2020} (\citeyear{Hahn2020}), which indicate that the BCF model is the most performing model out there for CATE estimation. The superiority of the BART-$(f_0, f_1)$ model over the BCF model also contradicts the conclusion of \citeauthor{BENCHMARKS2021_2a79ea27} (\citeyear{BENCHMARKS2021_2a79ea27}) related to the fact that indirect strategies for CATE estimation perform better than the direct alternative for the IHDP dataset. Additionally, the results of the IHDP experiments demonstrate that despite the promising results presented by the proposed model for the synthetic datasets, the K-Fold Causal BART model is likely to not be a state-of-the-art model that must be chosen over other alternatives for ATE and CATE estimation. Nevertheless, as \citeauthor{BENCHMARKS2021_2a79ea27} (\citeyear{BENCHMARKS2021_2a79ea27}) stated, IHDP is not 'one dataset' and only looking at the aggregated results of the experiments can be misleading. Instead, a percentile analysis based on the heterogeneity of the treatment $\tau(X)$ is more appropriate to draw proper conclusions of the experiments.

Yet, before moving to the percentile analysis, let's analyse the aggregated results of the ablation study using the IHDP dataset, which can be found in Table \ref{table6}. Starting with the first sub-model, Without $\mathbb{E}[\dot{\tau}(X)]$, it can be seen that the utilization of a regression of $\dot{\tau}(X)$ on $X$ anew improves the CATE estimation concerning all metrics. Moving to the second sub-model, Without Partialling Out and LM, it can be observed that  the use of the Double/Orthogonal machine learning method for the Parially Linear Model for ATE estimation decreases the accuracy of ATE estimation, which contradicts the results of the synthetic datasets. This results can be seen as evidence for the hypothesis that the IHDP dataset simply does not favor the use of the Double/Orthogonal machine learning method for the Partially Linear Model for ATE estimation. Considering the third sub-model, Without K-Fold for $\mathbb{E}[\dot{\tau}(X)]$, anew the employment of a second K-Fold method seems to not impact the model performance, thus adding unnecessary computational costs to it. Lastly, concerning the last sub-model, Without K-Fold for $\hat{Y}_1(X)$ and $\hat{Y}_0(X)$, the utilization of the K-Fold method for the T-Learner part greatly improves the ATE estimation accuracy while having a slightly negative impact on the CATE estimation. Thus, the results of the ablation study using the IHDP dataset give further evidence for the hypothesis that a second K-fold method is not necessary to further avoid overfitting and should be removed from the proposed model to make it more computationally efficient.

\begin{landscape}
    
\begin{table}[htbp]
\centering
\caption{Results of IHDP Experiments}
\begin{tabular}{>{\raggedright\arraybackslash}p{3cm} *{6}{c}}
\toprule
\textbf{Method} & \textbf{ATE RMSE} & \textbf{ATE Cover} & \textbf{ATE Length} & \textbf{CATE RMSE} & \textbf{CATE Cover} & \textbf{CATE Length} \\
\midrule
K-Fold Causal BART & 0.77 & 0.58 & 0.57  & 1.36  & 0.98 & 8.99 \\
BCF & 0.16 & 0.92 & 0.76 & 1.43 & 0.92 & 4.23 \\
ps-BART & 0.11  & 0.95 & 0.50 & 0.93  & 0.96  & 3.45 \\
BART-$(f_0, f_1)$ & 0.10 & 0.98 & 0.46 & 0.92  & 0.96  & 3.37 \\
Causal RF & 0.24 & 0.95 & 0.54 & 4.01 & 0.42 & 3.19 \\
LM + HS & 0.18 & 1.00 & 1.03  & 1.97 & 0.95  & 16.10  \\
\bottomrule
\end{tabular}
\label{table5}
\end{table}

\begin{table}[htbp]
\centering
\caption{Ablation Study Results of IHDP Experiments}
\begin{tabular}{>{\raggedright\arraybackslash}p{4.5cm} *{6}{c}}
\toprule
\textbf{Method} & \textbf{ATE RMSE} & \textbf{ATE Cover} & \textbf{ATE Length} & \textbf{CATE RMSE} & \textbf{CATE Cover} & \textbf{CATE Length} \\
\midrule
K-Fold Causal BART & 0.77  & 0.58 & 0.57 & 1.36  & 0.98 & 8.99 \\
Without $\mathbb{E}[\dot{\tau}(X)]$ & 0.77  & 0.58 & 0.57 & 1.50  & 0.99  & 11.19  \\
Without Partialling Out and LM & 0.10 & 0.29 & 0.11 & 1.36 & 0.98  & 8.99  \\
Without K-Fold for $\mathbb{E}[\dot{\tau}(X)]$ & 0.77 & 0.58 & 0.57 & 1.35  & 0.97  & 8.35  \\
Without K-Fold for $\hat{Y}_1(X)$ and $\hat{Y}_0(X)$ & 0.90  & 0.37 & 0.44 & 1.30 & 0.97  & 8.47 \\
\bottomrule
\end{tabular}
\label{table6}
\end{table}

\end{landscape}

\subsubsection{Percentile Analysis}

Table \ref{table:ihdp_cate_RMSE} presents the results of CATE RMSE per percentile of heterogeneity of the treatment $\tau(X)$. It is worth mentioning that the degree of heterogeneity of the treatment $\tau(X)$ present in a certain simulation of the IHDP is measured based on the relative standard deviation (SD), which is the standard deviation of the CATE values divided by the absolute value ATE (i.e., the standard deviation of $\tau(X)$ divided by the absolute mean of $\tau(X)$). It can be observed in Table \ref{table:ihdp_cate_RMSE} that up the 50th percentile, the most performing model is the BCF model, followed by the proposed model and the other BART models. Yet, after the 50th percentile, the BCF model and the K-Fold Causal BART model CATE estimation accuracy decreases to a greater extent than the other BART models. Hence, it can be concluded that if one believes that the dataset that they are handling in the context of CATE estimation presents a considerably high degree of heterogeneity of the treatment $\tau(X)$, then the BART-$(f_0,f_1)$ and the ps-BART models are preferred. Yet, in the converse situation, the BCF model would be preferred.

The results for ATE RMSE per percentile of heterogeneity can be found in Table \ref{table:ihdp_ate_RMSE}. Interestingly, the degree of heterogeneity of the treatment $\tau(X)$ does not seem to negatively impact the performance of the BART-$(f_0,f_1)$ and the ps-BART models, with the expection of the last decile. Although the performance of ATE estimation of the proposed model is below the other models for all deciles, it is in the last three deciles where this discrepancy becomes significantly big. This could indicate that the a high degree of heterogeneity of the treatment $\tau(X)$ does not favor the employment of  Double/Orthogonal machine learning method for the Partially Linear Model for ATE estimation, though this will only be verified in the quantile analysis of the ablation study. Finally, this time the BART-$(f_0,f_1)$ and the ps-BART models are preferred for all deciles.

Table \ref{table:ihdp_cate_cover} presents the CATE cover per percentile of heterogeneity results. Interestingly, it seems that most models are underconfident for the lower deciles, while yielding better CIs for the higher deciles (with the exception of the Causal RF and BCF models). Though in the aggregated results the cover of LM+HS model and the BART-$(f_0,f_1)$ and the ps-BART models are almost the same, it can be seen in Table \ref{table:ihdp_cate_cover} that the CIs constructed by the LM+HS model are much more consistent (i.e., close to 0.95 cover) than the other BART models.

The ATE cover per percentile of heterogeneity results can be seen in Table \ref{table:ihdp_ate_cover}. Different to the CATE results, there are no interesting patterns for the ATE results.

Table \ref{table:ihdp_cate_len} shows the results of CATE length per percentile of heterogeneity. Interestingly, until the last three deciles, the performance of the BCF model superior to the performance of the other models, especially for the first two quantiles while maintaining a proper cover in these deciles as seen in Table \ref{table:ihdp_cate_cover}. Consequently, if one believes that the dataset that they are handling in the context of CATE estimation presents a considerably low degree of heterogeneity of the treatment $\tau(X)$, then the BCF model is preferred not only due to its superiority in CATE estimation accuracy but also in uncertainty quantification. Lastly, as already seen in the aggregated results, the CATE length of the proposed model is systematically and considerably larger than the other models (with the exception of the LM+HS model), which is simply the result of the model's underconfidence for uncertainty quantification that could be a consequence of the heuristic method used to deal with the proposed model's systematical overconfidence. Hence, although the heuristic method used to deal with the proposed model's systematical overconfidence worked well for the synthetic datasets, it generalized poorly to the IHDP dataset, indicating that another approach should be taken to deal with the proposed model's systematical overconfidence.

The ATE length per percentile of heterogeneity results can be found in Table \ref{table:ihdp_ate_len}. Similarly to the ATE RMSE results, the ATE length results for the BART-$(f_0,f_1)$ and the ps-BART models seem to not suffer any impact from the degree of heterogeneity of the treatment $\tau(X)$. Thus, it could be affirmed that this demonstrates that the use of the BART-$(f_0,f_1)$ and the ps-BART models for ATE estimation is always preferred over other models independently of the degree of heterogeneity of the treatment $\tau(X)$ present in the handled dataset.

\begin{landscape}

\begin{table}[h!]
\centering
\caption{CATE RMSE per Percentile of Heterogeneity}
\begin{tabular}{|c|c|c|c|c|c|c|c|}
\hline
\textbf{Percentile} & \textbf{Relative SD} & \textbf{K-Fold Causal BART} & \textbf{Causal RF} & \textbf{BART-$(f_0,f_1)$} & \textbf{LM+HS} & \textbf{ps-BART} & \textbf{BCF} \\
\hline
0-10 & 0.06 & 0.35 & 0.27 & 0.39 & 0.26 & 0.39 & 0.26 \\
10-20 & 0.18 & 0.43 & 0.55 & 0.46 & 0.46 & 0.46 & 0.39 \\
20-30 & 0.27 & 0.48 & 0.85 & 0.49 & 0.60 & 0.49 & 0.49 \\
30-40 & 0.42 & 0.56 & 1.21 & 0.55 & 0.72 & 0.55 & 0.55 \\
40-50 & 0.55 & 0.67 & 1.61 & 0.64 & 0.96 & 0.64 & 0.68 \\
50-60 & 0.69 & 0.79 & 2.07 & 0.69 & 1.15 & 0.70 & 0.83 \\
60-70 & 0.88 & 0.77 & 2.43 & 0.67 & 1.20 & 0.68 & 0.78 \\
70-80 & 1.26 & 1.12 & 3.99 & 0.84 & 1.87 & 0.84 & 1.19 \\
80-90 & 2.18 & 3.86 & 12.72 & 2.29 & 6.93 & 2.28 & 4.17 \\
90-100 & 7.30 & 4.57 & 14.36 & 2.20 & 8.20 & 2.23 & 4.98 \\
\hline
\end{tabular}
\label{table:ihdp_cate_RMSE}
\end{table}

\begin{table}[h!]
\centering
\caption{ATE RMSE per Percentile of Heterogeneity}
\begin{tabular}{|c|c|c|c|c|c|c|c|}
\hline
\textbf{Percentile} & \textbf{Relative SD} & \textbf{K-Fold Causal BART} & \textbf{Causal RF} & \textbf{BART-$(f_0,f_1)$} & \textbf{LM+HS} & \textbf{ps-BART} & \textbf{BCF} \\
\hline
0-10 & 0.06 & 0.11 & 0.12 & 0.11 & 0.11 & 0.12 & 0.11 \\
10-20 & 0.18 & 0.08 & 0.06 & 0.06 & 0.03 & 0.05 & 0.05 \\
20-30 & 0.27 & 0.24 & 0.08 & 0.06 & 0.10 & 0.08 & 0.11 \\
30-40 & 0.42 & 0.18 & 0.10 & 0.09 & 0.12 & 0.09 & 0.09 \\
40-50 & 0.55 & 0.26 & 0.13 & 0.09 & 0.13 & 0.09 & 0.12 \\
50-60 & 0.69 & 0.31 & 0.16 & 0.10 & 0.11 & 0.11 & 0.10 \\
60-70 & 0.88 & 0.37 & 0.18 & 0.10 & 0.13 & 0.11 & 0.15 \\
70-80 & 1.26 & 0.81 & 0.15 & 0.08 & 0.12 & 0.08 & 0.14 \\
80-90 & 2.18 & 3.34 & 0.74 & 0.12 & 0.39 & 0.14 & 0.27 \\
90-100 & 7.30 & 1.98 & 0.65 & 0.17 & 0.55 & 0.21 & 0.47 \\
\hline
\end{tabular}
\label{table:ihdp_ate_RMSE}
\end{table}

\begin{table}[h!]
\centering
\caption{CATE Cover per Percentile of Heterogeneity}
\begin{tabular}{|c|c|c|c|c|c|c|c|}
\hline
\textbf{Percentile} & \textbf{Relative SD} & \textbf{K-Fold Causal BART} & \textbf{Causal RF} & \textbf{BART-$(f_0,f_1)$} & \textbf{LM+HS} & \textbf{ps-BART} & \textbf{BCF} \\
\hline
0-10 & 0.06 & 1.00 & 0.77 & 0.99 & 0.97 & 0.99 & 0.98 \\
10-20 & 0.18 & 1.00 & 0.59 & 0.99 & 0.96 & 0.99 & 0.96 \\
20-30 & 0.27 & 0.99 & 0.44 & 0.98 & 0.93 & 0.98 & 0.93 \\
30-40 & 0.42 & 0.98 & 0.41 & 0.98 & 0.95 & 0.98 & 0.95 \\
40-50 & 0.55 & 0.97 & 0.36 & 0.96 & 0.95 & 0.96 & 0.92 \\
50-60 & 0.69 & 0.96 & 0.35 & 0.96 & 0.94 & 0.96 & 0.91 \\
60-70 & 0.88 & 0.97 & 0.37 & 0.96 & 0.95 & 0.96 & 0.92 \\
70-80 & 1.26 & 0.97 & 0.33 & 0.95 & 0.95 & 0.96 & 0.92 \\
80-90 & 2.18 & 0.97 & 0.28 & 0.93 & 0.94 & 0.92 & 0.89 \\
90-100 & 7.30 & 0.97 & 0.34 & 0.94 & 0.95 & 0.94 & 0.87 \\
\hline
\end{tabular}
\label{table:ihdp_cate_cover}
\end{table}

\begin{table}[h!]
\centering
\caption{ATE Cover per Percentile of Heterogeneity}
\begin{tabular}{|c|c|c|c|c|c|c|c|}
\hline
\textbf{Percentile} & \textbf{Relative SD} & \textbf{K-Fold Causal BART} & \textbf{Causal RF} & \textbf{BART-$(f_0,f_1)$} & \textbf{LM+HS} & \textbf{ps-BART} & \textbf{BCF} \\
\hline
0-10 & 0.06 & 0.9 & 0.8 & 0.9 & 0.9 & 0.9 & 0.9 \\
10-20 & 0.18 & 0.9 & 1.0 & 1.0 & 1.0 & 1.0 & 1.0 \\
20-30 & 0.27 & 0.3 & 1.0 & 1.0 & 0.9 & 0.9 & 0.8 \\
30-40 & 0.42 & 0.6 & 1.0 & 1.0 & 1.0 & 1.0 & 1.0 \\
40-50 & 0.55 & 0.7 & 0.9 & 1.0 & 0.9 & 0.9 & 0.8 \\
50-60 & 0.69 & 0.5 & 0.9 & 1.0 & 0.9 & 1.0 & 0.9 \\
60-70 & 0.88 & 0.5 & 0.9 & 1.0 & 1.0 & 1.0 & 1.0 \\
70-80 & 1.26 & 0.4 & 1.0 & 0.9 & 1.0 & 0.9 & 0.9 \\
80-90 & 2.18 & 0.2 & 1.0 & 1.0 & 1.0 & 1.0 & 1.0 \\
90-100 & 7.30 & 0.8 & 1.0 & 1.0 & 1.0 & 0.9 & 0.9 \\
\hline
\end{tabular}
\label{table:ihdp_ate_cover}
\end{table}

\begin{table}[h!]
\centering
\caption{CATE Length per Percentile of Heterogeneity}
\begin{tabular}{|c|c|c|c|c|c|c|c|}
\hline
\textbf{Percentile} & \textbf{Relative SD} & \textbf{K-Fold Causal BART} & \textbf{Causal RF} & \textbf{BART-$(f_0,f_1)$} & \textbf{LM+HS} & \textbf{ps-BART} & \textbf{BCF} \\
\hline
0-10 & 0.06 & 4.32 & 0.57 & 2.15 & 1.39 & 2.19 & 1.36 \\
10-20 & 0.18 & 4.58 & 0.80 & 2.27 & 2.90 & 2.32 & 1.56 \\
20-30 & 0.27 & 4.70 & 0.95 & 2.31 & 3.86 & 2.37 & 1.68 \\
30-40 & 0.42 & 4.86 & 1.24 & 2.50 & 6.23 & 2.57 & 1.88 \\
40-50 & 0.55 & 5.25 & 1.51 & 2.65 & 8.04 & 2.73 & 2.23 \\
50-60 & 0.69 & 5.77 & 1.76 & 2.77 & 10.58 & 2.84 & 2.62 \\
60-70 & 0.88 & 5.93 & 2.16 & 2.78 & 13.09 & 2.85 & 2.51 \\
70-80 & 1.26 & 8.31 & 3.28 & 3.12 & 21.23 & 3.20 & 3.86 \\
80-90 & 2.18 & 22.73 & 9.11 & 6.24 & 60.89 & 6.44 & 11.92 \\
90-100 & 7.30 & 23.48 & 10.50 & 6.90 & 68.20 & 6.97 & 12.72 \\
\hline
\end{tabular}
\label{table:ihdp_cate_len}
\end{table}

\begin{table}[h!]
\centering
\caption{ATE Length per Percentile of Heterogeneity}
\begin{tabular}{|c|c|c|c|c|c|c|c|}
\hline
\textbf{Percentile} & \textbf{Relative SD} & \textbf{K-Fold Causal BART} & \textbf{Causal RF} & \textbf{BART-$(f_0,f_1)$} & \textbf{LM+HS} & \textbf{ps-BART} & \textbf{BCF} \\
\hline
0-10 & 0.06 & 0.20 & 0.19 & 0.43 & 0.42 & 0.47 & 0.43 \\
10-20 & 0.18 & 0.21 & 0.20 & 0.44 & 0.46 & 0.47 & 0.45 \\
20-30 & 0.27 & 0.22 & 0.21 & 0.42 & 0.48 & 0.46 & 0.46 \\
30-40 & 0.42 & 0.25 & 0.24 & 0.44 & 0.51 & 0.47 & 0.47 \\
40-50 & 0.55 & 0.28 & 0.27 & 0.44 & 0.57 & 0.47 & 0.50 \\
50-60 & 0.69 & 0.33 & 0.31 & 0.43 & 0.64 & 0.46 & 0.54 \\
60-70 & 0.88 & 0.36 & 0.35 & 0.43 & 0.67 & 0.47 & 0.53 \\
70-80 & 1.26 & 0.54 & 0.52 & 0.44 & 0.94 & 0.48 & 0.64 \\
80-90 & 2.18 & 1.58 & 1.43 & 0.56 & 2.90 & 0.59 & 1.69 \\
90-100 & 7.30 & 1.74 & 1.64 & 0.60 & 3.49 & 0.65 & 1.94 \\
\hline
\end{tabular}
\label{table:ihdp_ate_len}
\end{table}

\end{landscape}

Table \ref{table:ihdp_ablation_cate_RMSE} presents the results of CATE RMSE per percentile of heterogeneity for the ablation study. It can be anew observed that the use of a regression of $\dot{\tau}(X)$ on $X$ improves the CATE estimation accuracy for all deciles, though it does not considerably decrease the rate of performance loss for CATE estimation as the degree of heterogeneity of the treatment $\tau(X)$ increases. Considering the other sub-models, no considerable change in CATE estimation performance can be seen.

The ATE RMSE per percentile of heterogeneity results for the ablation study can be seen in Table \ref{table:ihdp_ablation_ate_RMSE}. Interestingly, the lack of a specialized approach for the ATE estimation (i.e., sub-model 2) greatly improves the ATE estimation precision and the decreases the rate of performance loss for ATE estimation as the degree of heterogeneity of the treatment $\tau(X)$ increases, making the ATE estimation precision even independent of the degree of heterogeneity. Nonetheless, given the results of the ablation study for the synthetic datasets, this is likely due to the fact that the IHDP dataset simply does not favor the use of the Double/Orthogonal machine learning method for the Partially Linear Model for ATE estimation. Additionally, as already seen in the aggregated results, the utilization of the K-Fold method for the T Learner part improves the ATE estimation accuracy for all deciles.

Table \ref{table:ihdp_ablation_cate_cover} shows the results of CATE cover per percentile of heterogeneity for the ablation study. It can be observed that the employment of only one K-Fold method instead of two actually improves the CATE cover, showing that the use of a second K-Fold method not only adds unnecessary computational costs to the model, but also decreases its uncertainty quantification precision to some extent.

The ATE cover per percentile of heterogeneity results for the ablation study can be found in Table \ref{table:ihdp_ablation_ate_cover}. The only two results worth mentioning are that the lack of a specialized approach for the ATE estimation (i.e., sub-model 2) and the lack of the K-Fold method for the estimation of $\hat{Y}_1(X)$ and $\hat{Y}_0(X)$ decreases ATE uncertainty quantification performance, which was already considerably poor.

Table \ref{table:ihdp_ablation_cate_len} presents the results of CATE length per percentile of heterogeneity for the ablation study. It can be anew observed that the use of a regression of $\dot{\tau}(X)$ on $X$ decreases the CATE length (and thus increasing its uncertainty quantification precision) for all deciles, though it does not considerably decrease the rate of performance loss for CATE uncertainty quantification as the degree of heterogeneity of the treatment $\tau(X)$ increases. Nevertheless, the uncertainty quantification precision is poor in both cases. Considering the other sub-models, no considerable change in CATE uncertainty quantification performance can be seen.

The results of ATE length per percentile of heterogeneity for the ablation study can be seen in Table \ref{table:ihdp_ablation_ate_len}. Though the lack of a specialized approach for the ATE estimation (i.e., sub-model 2) significantly decreases the ATE length for all deciles, such a decrease is meaningless given the fact that the ATE cover for this sub-model also considerably decreases. Considering the other sub-models, no considerable change in ATE uncertainty quantification performance can be seen.

\begin{landscape}

\begin{table}[h!]
\centering
\caption{CATE RMSE per Percentile of Heterogeneity (Ablation Study)}
\begin{tabular}{|c|c|c|c|c|c|c|}
\hline
\textbf{Percentile} & \textbf{Relative SD} & \textbf{K-Fold Causal BART} & \textbf{sub-model 1} & \textbf{sub-model 2} & \textbf{sub-model 3} & \textbf{sub-model 4} \\
\hline
0-10 & 0.06 & 0.35 & 0.41 & 0.35 & 0.35 & 0.38 \\
10-20 & 0.18 & 0.43 & 0.48 & 0.43 & 0.43 & 0.46 \\
20-30 & 0.27 & 0.48 & 0.53 & 0.48 & 0.47 & 0.49 \\
30-40 & 0.42 & 0.56 & 0.62 & 0.56 & 0.55 & 0.56 \\
40-50 & 0.55 & 0.67 & 0.76 & 0.67 & 0.67 & 0.67 \\
50-60 & 0.69 & 0.79 & 0.88 & 0.79 & 0.79 & 0.75 \\
60-70 & 0.88 & 0.77 & 0.85 & 0.77 & 0.76 & 0.72 \\
70-80 & 1.26 & 1.12 & 1.30 & 1.12 & 1.16 & 1.05 \\
80-90 & 2.18 & 3.86 & 4.23 & 3.86 & 3.86 & 3.65 \\
90-100 & 7.30 & 4.57 & 4.92 & 4.57 & 4.50 & 4.30 \\
\hline
\end{tabular}
\caption*{\textbf{sub-model 1}: Without $\mathbb{E}[\dot{\tau}(X)]$, \textbf{sub-model 2}: Without Partialling Out and LM, \textbf{sub-model 3}: Without K-Fold for $\mathbb{E}[\dot{\tau}(X)]$, \textbf{sub-model 4}: Without K-Fold for $\hat{Y}_1(X)$ and $\hat{Y}_0(X)$.}
\label{table:ihdp_ablation_cate_RMSE}
\end{table}

\begin{table}[h!]
\centering
\caption{ATE RMSE per Percentile of Heterogeneity (Ablation Study)}
\begin{tabular}{|c|c|c|c|c|c|c|}
\hline
\textbf{Percentile} & \textbf{Relative SD} & \textbf{K-Fold Causal BART} & \textbf{sub-model 1} & \textbf{sub-model 2} & \textbf{sub-model 3} & \textbf{sub-model 4} \\
\hline
0-10 & 0.06 & 0.11 & 0.11 & 0.11 & 0.11 & 0.15 \\
10-20 & 0.18 & 0.08 & 0.08 & 0.06 & 0.08 & 0.13 \\
20-30 & 0.27 & 0.24 & 0.24 & 0.06 & 0.24 & 0.31 \\
30-40 & 0.42 & 0.18 & 0.18 & 0.09 & 0.18 & 0.31 \\
40-50 & 0.55 & 0.26 & 0.26 & 0.09 & 0.26 & 0.46 \\
50-60 & 0.69 & 0.31 & 0.31 & 0.10 & 0.31 & 0.63 \\
60-70 & 0.88 & 0.37 & 0.37 & 0.10 & 0.37 & 0.55 \\
70-80 & 1.26 & 0.81 & 0.81 & 0.09 & 0.81 & 1.11 \\
80-90 & 2.18 & 3.34 & 3.34 & 0.12 & 3.34 & 3.53 \\
90-100 & 7.30 & 1.98 & 1.98 & 0.17 & 1.98 & 1.80 \\
\hline
\end{tabular}
\caption*{\textbf{sub-model 1}: Without $\mathbb{E}[\dot{\tau}(X)]$, \textbf{sub-model 2}: Without Partialling Out and LM, \textbf{sub-model 3}: Without K-Fold for $\mathbb{E}[\dot{\tau}(X)]$, \textbf{sub-model 4}: Without K-Fold for $\hat{Y}_1(X)$ and $\hat{Y}_0(X)$.}
\label{table:ihdp_ablation_ate_RMSE}
\end{table}

\begin{table}[h!]
\centering
\caption{CATE Cover per Percentile of Heterogeneity (Ablation Study)}
\begin{tabular}{|c|c|c|c|c|c|c|}
\hline
\textbf{Percentile} & \textbf{Relative SD} & \textbf{K-Fold Causal BART} & \textbf{sub-model 1} & \textbf{sub-model 2} & \textbf{sub-model 3} & \textbf{sub-model 4} \\
\hline
0-10 & 0.06 & 1.00 & 1.00 & 1.00 & 1.00 & 1.00 \\
10-20 & 0.18 & 1.00 & 1.00 & 1.00 & 1.00 & 0.99 \\
20-30 & 0.27 & 0.99 & 1.00 & 0.99 & 0.99 & 0.98 \\
30-40 & 0.42 & 0.98 & 1.00 & 0.98 & 0.97 & 0.96 \\
40-50 & 0.55 & 0.97 & 1.00 & 0.97 & 0.96 & 0.95 \\
50-60 & 0.69 & 0.96 & 1.00 & 0.96 & 0.95 & 0.95 \\
60-70 & 0.88 & 0.97 & 1.00 & 0.97 & 0.95 & 0.95 \\
70-80 & 1.26 & 0.97 & 0.99 & 0.97 & 0.96 & 0.96 \\
80-90 & 2.18 & 0.97 & 0.98 & 0.97 & 0.96 & 0.97 \\
90-100 & 7.30 & 0.97 & 0.98 & 0.97 & 0.96 & 0.96 \\
\hline
\end{tabular}
\caption*{\textbf{sub-model 1}: Without $\mathbb{E}[\dot{\tau}(X)]$, \textbf{sub-model 2}: Without Partialling Out and LM, \textbf{sub-model 3}: Without K-Fold for $\mathbb{E}[\dot{\tau}(X)]$, \textbf{sub-model 4}: Without K-Fold for $\hat{Y}_1(X)$ and $\hat{Y}_0(X)$.}
\label{table:ihdp_ablation_cate_cover}
\end{table}

\begin{table}[h!]
\centering
\caption{ATE Cover per Percentile of Heterogeneity (Ablation Study)}
\begin{tabular}{|c|c|c|c|c|c|c|}
\hline
\textbf{Percentile} & \textbf{Relative SD} & \textbf{K-Fold Causal BART} & \textbf{sub-model 1} & \textbf{sub-model 2} & \textbf{sub-model 3} & \textbf{sub-model 4} \\
\hline
0-10 & 0.06 & 0.9 & 0.9 & 0.1 & 0.9 & 0.7 \\
10-20 & 0.18 & 0.9 & 0.9 & 0.1 & 0.9 & 0.8 \\
20-30 & 0.27 & 0.3 & 0.3 & 0.3 & 0.3 & 0.3 \\
30-40 & 0.42 & 0.6 & 0.6 & 0.1 & 0.6 & 0.4 \\
40-50 & 0.55 & 0.7 & 0.7 & 0.2 & 0.7 & 0.3 \\
50-60 & 0.69 & 0.5 & 0.5 & 0.2 & 0.5 & 0.1 \\
60-70 & 0.88 & 0.5 & 0.5 & 0.2 & 0.5 & 0.3 \\
70-80 & 1.26 & 0.4 & 0.4 & 0.3 & 0.4 & 0.3 \\
80-90 & 2.18 & 0.2 & 0.2 & 0.9 & 0.2 & 0.1 \\
90-100 & 7.30 & 0.8 & 0.8 & 0.5 & 0.8 & 0.4 \\
\hline
\end{tabular}
\caption*{\textbf{sub-model 1}: Without $\mathbb{E}[\dot{\tau}(X)]$, \textbf{sub-model 2}: Without Partialling Out and LM, \textbf{sub-model 3}: Without K-Fold for $\mathbb{E}[\dot{\tau}(X)]$, \textbf{sub-model 4}: Without K-Fold for $\hat{Y}_1(X)$ and $\hat{Y}_0(X)$.}
\label{table:ihdp_ablation_ate_cover}
\end{table}

\begin{table}[h!]
\centering
\caption{CATE Length per Percentile of Heterogeneity (Ablation Study)}
\begin{tabular}{|c|c|c|c|c|c|c|}
\hline
\textbf{Percentile} & \textbf{Relative SD} & \textbf{K-Fold Causal BART} & \textbf{sub-model 1} & \textbf{sub-model 2} & \textbf{sub-model 3} & \textbf{sub-model 4} \\
\hline
0-10 & 0.06 & 4.32 & 6.26 & 4.32 & 4.27 & 4.06 \\
10-20 & 0.18 & 4.58 & 6.55 & 4.58 & 4.48 & 4.24 \\
20-30 & 0.27 & 4.70 & 6.69 & 4.70 & 4.58 & 4.34 \\
30-40 & 0.42 & 4.86 & 7.01 & 4.86 & 4.69 & 4.47 \\
40-50 & 0.55 & 5.25 & 7.47 & 5.25 & 5.02 & 4.85 \\
50-60 & 0.69 & 5.77 & 8.02 & 5.77 & 5.47 & 5.41 \\
60-70 & 0.88 & 5.93 & 8.22 & 5.93 & 5.61 & 5.51 \\
70-80 & 1.26 & 8.31 & 10.53 & 8.31 & 7.78 & 7.71 \\
80-90 & 2.18 & 22.73 & 25.08 & 22.73 & 20.66 & 21.85 \\
90-100 & 7.30 & 23.48 & 26.09 & 23.48 & 20.92 & 22.26 \\
\hline
\end{tabular}
\caption*{\textbf{sub-model 1}: Without $\mathbb{E}[\dot{\tau}(X)]$, \textbf{sub-model 2}: Without Partialling Out and LM, \textbf{sub-model 3}: Without K-Fold for $\mathbb{E}[\dot{\tau}(X)]$, \textbf{sub-model 4}: Without K-Fold for $\hat{Y}_1(X)$ and $\hat{Y}_0(X)$.}
\label{table:ihdp_ablation_cate_len}
\end{table}

\begin{table}[h!]
\centering
\caption{ATE Length per Percentile of Heterogeneity (Ablation Study)}
\begin{tabular}{|c|c|c|c|c|c|c|}
\hline
\textbf{Percentile} & \textbf{Relative SD} & \textbf{K-Fold Causal BART} & \textbf{sub-model 1} & \textbf{sub-model 2} & \textbf{sub-model 3} & \textbf{sub-model 4} \\
\hline
0-10 & 0.06 & 0.20 & 0.20 & 0.02 & 0.20 & 0.19 \\
10-20 & 0.18 & 0.21 & 0.21 & 0.03 & 0.21 & 0.20 \\
20-30 & 0.27 & 0.22 & 0.22 & 0.03 & 0.22 & 0.21 \\
30-40 & 0.42 & 0.25 & 0.25 & 0.04 & 0.25 & 0.23 \\
40-50 & 0.55 & 0.28 & 0.28 & 0.05 & 0.28 & 0.25 \\
50-60 & 0.69 & 0.33 & 0.33 & 0.06 & 0.33 & 0.29 \\
60-70 & 0.88 & 0.36 & 0.36 & 0.06 & 0.36 & 0.31 \\
70-80 & 1.26 & 0.54 & 0.54 & 0.09 & 0.54 & 0.43 \\
80-90 & 2.18 & 1.58 & 1.58 & 0.34 & 1.58 & 1.18 \\
90-100 & 7.30 & 1.74 & 1.74 & 0.39 & 1.74 & 1.14 \\
\hline
\end{tabular}
\caption*{\textbf{sub-model 1}: Without $\mathbb{E}[\dot{\tau}(X)]$, \textbf{sub-model 2}: Without Partialling Out and LM, \textbf{sub-model 3}: Without K-Fold for $\mathbb{E}[\dot{\tau}(X)]$, \textbf{sub-model 4}: Without K-Fold for $\hat{Y}_1(X)$ and $\hat{Y}_0(X)$.}
\label{table:ihdp_ablation_ate_len}
\end{table}

\end{landscape}

\section{Conclusion and Novel Insights}\label{Conclusion and Novel Insights}

The primary aim of this research was to propose a novel model, the K-Fold Causal BART, to improve the estimation of ATE and CATE. Despite the promising results demonstrated by the proposed model for synthetic datasets, the findings from the IHDP dataset suggest that the K-Fold Causal BART is not a state-of-the-art model for CATE and ATE estimation. Nevertheless, this research provides several novel and important insights that can significantly impact the current literature on CATE estimation.

Firstly, the performance of the ps-BART model across both synthetic and IHDP datasets indicates that it is likely the preferred model for CATE and ATE estimation, demonstrating better generalization capabilities than the BCF model. This challenges the current perception in the literature that the BCF model is the best for CATE estimation.

Secondly, the research reveals that while the BCF model performs better than other benchmark models in scenarios with relatively low heterogeneity of the treatment effect, its performance significantly deteriorates with increased heterogeneity. Conversely, the ps-BART and BART-$(f_0, f_1)$ models perform better than other benchmark models under high heterogeneity conditions. Therefore, in the presence of low treatment effect heterogeneity, the BCF model is preferable due to its superior accuracy and uncertainty quantification for CATE estimation, while in the presence of high treatment effect heterogeneity, the ps-BART and BART-$(f_0, f_1)$ models are preferable. However, for ATE estimation, the ps-BART model consistently outperforms others regardless of the treatment effect heterogeneity.

Additionally, this study highlights that causal inference models tend to be overconfident in their CATE uncertainty quantification when the treatment effect heterogeneity is low. Future research should explore methods to address this overconfidence to improve model reliability and also propose theoretical explanations for this phenomenon. Another significant insight is the redundancy of employing a second K-Fold method to avoid model overfitting for CATE and ATE estimation. The results show that a single K-Fold method is sufficient, with the second K-Fold method not only adding unnecessary computational costs but also reducing the precision of uncertainty quantification.

Furthermore, the study underscores the importance of thoroughly understanding the datasets used for causal inference models. It demonstrates that a more nuanced analysis than simple aggregated results is essential for gaining insights into model performance and how specific dataset characteristics influence these performances. This approach can provide deeper insights into both novel and benchmark models, guiding future research directions.

Lastly, the superiority of the BART-$(f_0, f_1)$ model over the BCF model contradicts previous conclusions by \citeauthor{BENCHMARKS2021_2a79ea27} (\citeyear{BENCHMARKS2021_2a79ea27}) that indirect strategies for CATE estimation are preferable for the IHDP dataset. This contradiction highlights the need for continual evaluation and comparison of causal inference models to refine our understanding of their capabilities and limitations.

In conclusion, while the proposed K-Fold Causal BART model may not emerge as the new standard for CATE and ATE estimation, the insights gained from this research provide valuable contributions to the field. These findings encourage a reevaluation of current best practices and promote the development of more robust and adaptable causal inference methodologies.

\section{Research Limitations}\label{Research Limitations}

Despite the significant insights and contributions presented by this research, several limitations must be acknowledged to provide a balanced perspective on the findings and to guide future work in this domain.

The use of synthetic DGPs inherently introduces certain biases. While these DGPs are designed to simulate a wide range of treatment effects and covariate distributions, they are still simplifications of real-world complexities \parencite{Hahn2020,BENCHMARKS2021_2a79ea27}. The assumptions made in the DGPs, such as specific functional forms and distributions, might not fully capture the intricacies present in actual observational data \parencite{Hahn2020,BENCHMARKS2021_2a79ea27}. Consequently, the performance of the K-Fold Causal BART model and other benchmark models might differ when applied to real-world datasets with unanticipated complexities and noise.

Additionally, the IHDP benchmark dataset, despite its widespread use and recognition, also has its limitations \parencite{BENCHMARKS2021_2a79ea27}. The dataset is derived from a controlled experiment with added confounding to simulate observational study conditions. However, the introduced confounding may not perfectly replicate the complexities and nuances of real-world observational data \parencite{BENCHMARKS2021_2a79ea27}. Additionally, the specific modifications made to the dataset and treatment relationship with the outcome variable can favor certain models over others \parencite{BENCHMARKS2021_2a79ea27}.

While the proposed K-Fold Causal BART model demonstrated competitive performance in synthetic datasets, its generalization to real-world scenarios remains uncertain. The discrepancies observed in the IHDP dataset suggest that the model might not be universally applicable across different types of data. Similarly, the fact that the ps-BART model generalized better than the BCF model for the IHDP dataset might not necessarily mean that the ps-BART model inherently and always generalizes better than the BCF model. The specific nature of the IHDP dataset could have favored the ps-BART model over the BCF model.

Incidentally, the selection of benchmark models was inspired by \citeauthor{Hahn2020} (\citeyear{Hahn2020}) and included a range of popular models in the literature. However, the conclusions drawn from comparative analyses are limited by the specific models chosen. Other potential benchmark models, which were not included in this study, might offer different insights and performance characteristics. Future research should consider a broader range of models to ensure a comprehensive evaluation. Besides, the ablation study provided valuable insights into the components of the K-Fold Causal BART model, yet it also highlighted certain constraints. The study's design focused on specific sub-models and their impacts on performance metrics. However, the scope of the ablation study was limited to particular methodological aspects, and other potential modifications or components were not explored. Further research could expand the ablation study to include additional factors and interactions that might influence model performance.

Lastly, the results indicated that the performance of causal inference models, including the proposed K-Fold Causal BART model, can be highly dataset-specific, as already pointed out by \citeauthor{BENCHMARKS2021_2a79ea27} (\citeyear{BENCHMARKS2021_2a79ea27}). While the model performed better than the benchmark models in synthetic scenarios, its performance in the IHDP dataset revealed limitations. This highlights the necessity for future research to conduct extensive validation across multiple, diverse datasets to ensure the robustness and generalizability of the findings.

Given these limitations, future research should focus on several key areas:
\begin{itemize}
    \item Expanding the range of synthetic and real-world datasets to test model robustness.
    \item Exploring alternative methodological approaches to improve model generalization and computational efficiency.
    \item Enhancing uncertainty quantification methods to address overconfidence and underconfidence issues.
    \item Conducting broader ablation studies to investigate additional model components and their interactions.
    \item Validating findings across diverse datasets to ensure applicability and reliability in varied contexts.
\end{itemize}

In summary, while this research has made significant contributions to the field of CATE estimation, recognizing and addressing these limitations is essential for advancing the development and application of robust causal inference models.

\printbibliography

@inproceedings{NEURIPS2022_0378c769,
 author = {Grinsztajn, Leo and Oyallon, Edouard and Varoquaux, Gael},
 booktitle = {Advances in Neural Information Processing Systems},
 editor = {S. Koyejo and S. Mohamed and A. Agarwal and D. Belgrave and K. Cho and A. Oh},
 pages = {507--520},
 publisher = {Curran Associates, Inc.},
 title = {Why do tree-based models still outperform deep learning on typical tabular data?},
 url = {https://proceedings.neurips.cc/paper_files/paper/2022/file/0378c7692da36807bdec87ab043cdadc-Paper-Datasets_and_Benchmarks.pdf},
 volume = {35},
 year = {2022}
}

@InProceedings{pmlr-v130-curth21a,
  title = 	 { Nonparametric Estimation of Heterogeneous Treatment Effects: From Theory to Learning Algorithms },
  author =       {Curth, Alicia and van der Schaar, Mihaela},
  booktitle = 	 {Proceedings of The 24th International Conference on Artificial Intelligence and Statistics},
  pages = 	 {1810--1818},
  year = 	 {2021},
  editor = 	 {Banerjee, Arindam and Fukumizu, Kenji},
  volume = 	 {130},
  series = 	 {Proceedings of Machine Learning Research},
  month = 	 {13--15 Apr},
  publisher =    {PMLR},
  url = 	 {https://proceedings.mlr.press/v130/curth21a.html},
}

@inproceedings{BENCHMARKS2021_2a79ea27,
 author = {Curth, Alicia and Svensson, David and Weatherall, Jim and van der Schaar, Mihaela},
 booktitle = {Proceedings of the Neural Information Processing Systems Track on Datasets and Benchmarks},
 editor = {J. Vanschoren and S. Yeung},
 pages = {},
 publisher = {Curran},
 title = {Really Doing Great at Estimating CATE? A Critical Look at ML Benchmarking Practices in Treatment Effect Estimation},
 url = {https://datasets-benchmarks-proceedings.neurips.cc/paper_files/paper/2021/file/2a79ea27c279e471f4d180b08d62b00a-Paper-round2.pdf},
 volume = {1},
 year = {2021}
}

@article{Hahn2020,
  title = {Bayesian Regression Tree Models for Causal Inference: Regularization,  Confounding,  and Heterogeneous Effects (with Discussion)},
  volume = {15},
  ISSN = {1936-0975},
  url = {http://dx.doi.org/10.1214/19-BA1195},
  DOI = {10.1214/19-ba1195},
  number = {3},
  journal = {Bayesian Analysis},
  publisher = {Institute of Mathematical Statistics},
  author = {Hahn,  P. Richard and Murray,  Jared S. and Carvalho,  Carlos M.},
  year = {2020},
  month = sep 
}

@article{10.1145/3444944,
author = {Yao, Liuyi and Chu, Zhixuan and Li, Sheng and Li, Yaliang and Gao, Jing and Zhang, Aidong},
title = {A Survey on Causal Inference},
year = {2021},
issue_date = {October 2021},
publisher = {Association for Computing Machinery},
address = {New York, NY, USA},
volume = {15},
number = {5},
issn = {1556-4681},
url = {https://doi.org/10.1145/3444944},
doi = {10.1145/3444944},
journal = {ACM Trans. Knowl. Discov. Data},
month = {may},
articleno = {74},
numpages = {46}
}

@inproceedings{10.1145/3394486.3406460,
author = {Cui, Peng and Shen, Zheyan and Li, Sheng and Yao, Liuyi and Li, Yaliang and Chu, Zhixuan and Gao, Jing},
title = {Causal Inference Meets Machine Learning},
year = {2020},
isbn = {9781450379984},
publisher = {Association for Computing Machinery},
address = {New York, NY, USA},
url = {https://doi.org/10.1145/3394486.3406460},
doi = {10.1145/3394486.3406460},
pages = {3527–3528},
numpages = {2},
location = {Virtual Event, CA, USA},
series = {KDD '20}
}

@article{Kuang2020,
  title = {Causal Inference},
  volume = {6},
  ISSN = {2095-8099},
  url = {http://dx.doi.org/10.1016/j.eng.2019.08.016},
  DOI = {10.1016/j.eng.2019.08.016},
  number = {3},
  journal = {Engineering},
  publisher = {Elsevier BV},
  author = {Kuang,  Kun and Li,  Lian and Geng,  Zhi and Xu,  Lei and Zhang,  Kun and Liao,  Beishui and Huang,  Huaxin and Ding,  Peng and Miao,  Wang and Jiang,  Zhichao},
  year = {2020},
  month = mar,
  pages = {253–263}
}

@article{Crown2019,
  title = {Real-World Evidence,  Causal Inference,  and Machine Learning},
  volume = {22},
  ISSN = {1098-3015},
  url = {http://dx.doi.org/10.1016/j.jval.2019.03.001},
  DOI = {10.1016/j.jval.2019.03.001},
  number = {5},
  journal = {Value in Health},
  publisher = {Elsevier BV},
  author = {Crown,  William H.},
  year = {2019},
  month = may,
  pages = {587–592}
}

@article{Grimmer2014,
  title = {We Are All Social Scientists Now: How Big Data,  Machine Learning,  and Causal Inference Work Together},
  volume = {48},
  ISSN = {1537-5935},
  url = {http://dx.doi.org/10.1017/S1049096514001784},
  DOI = {10.1017/s1049096514001784},
  number = {01},
  journal = {PS: Political Science \& Politics},
  publisher = {Cambridge University Press (CUP)},
  author = {Grimmer,  Justin},
  year = {2014},
  month = dec,
  pages = {80–83}
}

@article{Ohlsson2020,
  title = {Applying Causal Inference Methods in Psychiatric Epidemiology: A Review},
  volume = {77},
  ISSN = {2168-622X},
  url = {http://dx.doi.org/10.1001/jamapsychiatry.2019.3758},
  DOI = {10.1001/jamapsychiatry.2019.3758},
  number = {6},
  journal = {JAMA Psychiatry},
  publisher = {American Medical Association (AMA)},
  author = {Ohlsson,  Henrik and Kendler,  Kenneth S.},
  year = {2020},
  month = jun,
  pages = {637}
}

@article{Bembom2007,
  title = {A practical illustration of the importance of realistic individualized treatment rules in causal inference},
  volume = {1},
  ISSN = {1935-7524},
  url = {http://dx.doi.org/10.1214/07-EJS105},
  DOI = {10.1214/07-ejs105},
  number = {none},
  journal = {Electronic Journal of Statistics},
  publisher = {Institute of Mathematical Statistics},
  author = {Bembom,  Oliver and van der Laan,  Mark J.},
  year = {2007},
  month = jan 
}

@article{Sobel2000,
  title = {Causal Inference in the Social Sciences},
  volume = {95},
  ISSN = {1537-274X},
  url = {http://dx.doi.org/10.1080/01621459.2000.10474243},
  DOI = {10.1080/01621459.2000.10474243},
  number = {450},
  journal = {Journal of the American Statistical Association},
  publisher = {Informa UK Limited},
  author = {Sobel,  Michael E.},
  year = {2000},
  month = jun,
  pages = {647–651}
}

@book{murnane2010methods,
  title={Methods matter: Improving causal inference in educational and social science research},
  author={Murnane, Richard J and Willett, John B},
  year={2010},
  publisher={Oxford University Press}
}

@article{Pan2018,
  title = {Propensity score methods for causal inference: an overview},
  volume = {45},
  ISSN = {1349-6964},
  url = {http://dx.doi.org/10.1007/s41237-018-0058-8},
  DOI = {10.1007/s41237-018-0058-8},
  number = {2},
  journal = {Behaviormetrika},
  publisher = {Springer Science and Business Media LLC},
  author = {Pan,  Wei and Bai,  Haiyan},
  year = {2018},
  month = jul,
  pages = {317–334}
}

@article{Shiba2021,
  title = {Using Propensity Scores for Causal Inference: Pitfalls and Tips},
  volume = {31},
  ISSN = {1349-9092},
  url = {http://dx.doi.org/10.2188/jea.JE20210145},
  DOI = {10.2188/jea.je20210145},
  number = {8},
  journal = {Journal of Epidemiology},
  publisher = {Japan Epidemiological Association},
  author = {Shiba,  Koichiro and Kawahara,  Takuya},
  year = {2021},
  month = aug,
  pages = {457–463}
}

@article{Imai2004,
  title = {Causal Inference With General Treatment Regimes: Generalizing the Propensity Score},
  volume = {99},
  ISSN = {1537-274X},
  url = {http://dx.doi.org/10.1198/016214504000001187},
  DOI = {10.1198/016214504000001187},
  number = {467},
  journal = {Journal of the American Statistical Association},
  publisher = {Informa UK Limited},
  author = {Imai,  Kosuke and van Dyk,  David A},
  year = {2004},
  month = sep,
  pages = {854–866}
}

@article{Tan2006,
  title = {A Distributional Approach for Causal Inference Using Propensity Scores},
  volume = {101},
  ISSN = {1537-274X},
  url = {http://dx.doi.org/10.1198/016214506000000023},
  DOI = {10.1198/016214506000000023},
  number = {476},
  journal = {Journal of the American Statistical Association},
  publisher = {Informa UK Limited},
  author = {Tan,  Zhiqiang},
  year = {2006},
  month = dec,
  pages = {1619–1637}
}

@article{Li2012,
  title = {Using the Propensity Score Method to Estimate Causal Effects: A Review and Practical Guide},
  volume = {16},
  ISSN = {1552-7425},
  url = {http://dx.doi.org/10.1177/1094428112447816},
  DOI = {10.1177/1094428112447816},
  number = {2},
  journal = {Organizational Research Methods},
  publisher = {SAGE Publications},
  author = {Li,  Mingxiang},
  year = {2012},
  month = jun,
  pages = {188–226}
}

@article{Fuentes2021,
  title = {Causal Inference with Multilevel Data: A Comparison of Different Propensity Score Weighting Approaches},
  volume = {57},
  ISSN = {1532-7906},
  url = {http://dx.doi.org/10.1080/00273171.2021.1925521},
  DOI = {10.1080/00273171.2021.1925521},
  number = {6},
  journal = {Multivariate Behavioral Research},
  publisher = {Informa UK Limited},
  author = {Fuentes,  Alvaro and L\"{u}dtke,  Oliver and Robitzsch,  Alexander},
  year = {2021},
  month = jun,
  pages = {916–939}
}

@misc{https://doi.org/10.48550/arxiv.2004.14497,
  doi = {10.48550/ARXIV.2004.14497},
  url = {https://arxiv.org/abs/2004.14497},
  author = {Kennedy,  Edward H.},
  title = {Towards optimal doubly robust estimation of heterogeneous causal effects},
  publisher = {arXiv},
  year = {2020},
  copyright = {arXiv.org perpetual,  non-exclusive license}
}

@article{Knzel2019,
  title = {Metalearners for estimating heterogeneous treatment effects using machine learning},
  volume = {116},
  ISSN = {1091-6490},
  url = {http://dx.doi.org/10.1073/pnas.1804597116},
  DOI = {10.1073/pnas.1804597116},
  number = {10},
  journal = {Proceedings of the National Academy of Sciences},
  publisher = {Proceedings of the National Academy of Sciences},
  author = {K\"{u}nzel,  S\"{o}ren R. and Sekhon,  Jasjeet S. and Bickel,  Peter J. and Yu,  Bin},
  year = {2019},
  month = feb,
  pages = {4156–4165}
}

@article{Wager2018,
  title = {Estimation and Inference of Heterogeneous Treatment Effects using Random Forests},
  volume = {113},
  ISSN = {1537-274X},
  url = {http://dx.doi.org/10.1080/01621459.2017.1319839},
  DOI = {10.1080/01621459.2017.1319839},
  number = {523},
  journal = {Journal of the American Statistical Association},
  publisher = {Informa UK Limited},
  author = {Wager,  Stefan and Athey,  Susan},
  year = {2018},
  month = jun,
  pages = {1228–1242}
}

@InProceedings{pmlr-v97-oprescu19a,
  title = 	 {Orthogonal Random Forest for Causal Inference},
  author =       {Oprescu, Miruna and Syrgkanis, Vasilis and Wu, Zhiwei Steven},
  booktitle = 	 {Proceedings of the 36th International Conference on Machine Learning},
  pages = 	 {4932--4941},
  year = 	 {2019},
  editor = 	 {Chaudhuri, Kamalika and Salakhutdinov, Ruslan},
  volume = 	 {97},
  series = 	 {Proceedings of Machine Learning Research},
  month = 	 {09--15 Jun},
  publisher =    {PMLR},
  url = 	 {https://proceedings.mlr.press/v97/oprescu19a.html},
}

@InProceedings{pmlr-v70-shalit17a,
  title = 	 {Estimating individual treatment effect: generalization bounds and algorithms},
  author =       {Uri Shalit and Fredrik D. Johansson and David Sontag},
  booktitle = 	 {Proceedings of the 34th International Conference on Machine Learning},
  pages = 	 {3076--3085},
  year = 	 {2017},
  editor = 	 {Precup, Doina and Teh, Yee Whye},
  volume = 	 {70},
  series = 	 {Proceedings of Machine Learning Research},
  month = 	 {06--11 Aug},
  publisher =    {PMLR},
  url = 	 {https://proceedings.mlr.press/v70/shalit17a.html},
}

@inproceedings{NEURIPS2019_8fb5f8be,
 author = {Shi, Claudia and Blei, David and Veitch, Victor},
 booktitle = {Advances in Neural Information Processing Systems},
 editor = {H. Wallach and H. Larochelle and A. Beygelzimer and F. d\textquotesingle Alch\'{e}-Buc and E. Fox and R. Garnett},
 pages = {},
 publisher = {Curran Associates, Inc.},
 title = {Adapting Neural Networks for the Estimation of Treatment Effects},
 url = {https://proceedings.neurips.cc/paper_files/paper/2019/file/8fb5f8be2aa9d6c64a04e3ab9f63feee-Paper.pdf},
 volume = {32},
 year = {2019}
}

@inproceedings{
Hassanpour2020Learning,
title={Learning Disentangled Representations for CounterFactual Regression},
author={Negar Hassanpour and Russell Greiner},
booktitle={International Conference on Learning Representations},
year={2020},
url={https://openreview.net/forum?id=HkxBJT4YvB}
}

@article{Hahn2018,
  title = {Regularization and Confounding in Linear Regression for Treatment Effect Estimation},
  volume = {13},
  ISSN = {1936-0975},
  url = {http://dx.doi.org/10.1214/16-BA1044},
  DOI = {10.1214/16-ba1044},
  number = {1},
  journal = {Bayesian Analysis},
  publisher = {Institute of Mathematical Statistics},
  author = {Hahn,  P. Richard and Carvalho,  Carlos M. and Puelz,  David and He,  Jingyu},
  year = {2018},
  month = mar 
}

@article{Yeager2019,
  title = {A national experiment reveals where a growth mindset improves achievement},
  volume = {573},
  ISSN = {1476-4687},
  url = {http://dx.doi.org/10.1038/s41586-019-1466-y},
  DOI = {10.1038/s41586-019-1466-y},
  number = {7774},
  journal = {Nature},
  publisher = {Springer Science and Business Media LLC},
  author = {Yeager,  David S. and Hanselman,  Paul and Walton,  Gregory M. and Murray,  Jared S. and Crosnoe,  Robert and Muller,  Chandra and Tipton,  Elizabeth and Schneider,  Barbara and Hulleman,  Chris S. and Hinojosa,  Cintia P. and Paunesku,  David and Romero,  Carissa and Flint,  Kate and Roberts,  Alice and Trott,  Jill and Iachan,  Ronaldo and Buontempo,  Jenny and Yang,  Sophia Man and Carvalho,  Carlos M. and Hahn,  P. Richard and Gopalan,  Maithreyi and Mhatre,  Pratik and Ferguson,  Ronald and Duckworth,  Angela L. and Dweck,  Carol S.},
  year = {2019},
  month = aug,
  pages = {364–369}
}

@article{Bail2019,
  title = {Assessing the Russian Internet Research Agency’s impact on the political attitudes and behaviors of American Twitter users in late 2017},
  volume = {117},
  ISSN = {1091-6490},
  url = {http://dx.doi.org/10.1073/pnas.1906420116},
  DOI = {10.1073/pnas.1906420116},
  number = {1},
  journal = {Proceedings of the National Academy of Sciences},
  publisher = {Proceedings of the National Academy of Sciences},
  author = {Bail,  Christopher A. and Guay,  Brian and Maloney,  Emily and Combs,  Aidan and Hillygus,  D. Sunshine and Merhout,  Friedolin and Freelon,  Deen and Volfovsky,  Alexander},
  year = {2019},
  month = nov,
  pages = {243–250}
}

@article{Yeager2022,
  title = {A synergistic mindsets intervention protects adolescents from stress},
  volume = {607},
  ISSN = {1476-4687},
  url = {http://dx.doi.org/10.1038/s41586-022-04907-7},
  DOI = {10.1038/s41586-022-04907-7},
  number = {7919},
  journal = {Nature},
  publisher = {Springer Science and Business Media LLC},
  author = {Yeager,  David S. and Bryan,  Christopher J. and Gross,  James J. and Murray,  Jared S. and Krettek Cobb,  Danielle and H. F. Santos,  Pedro and Gravelding,  Hannah and Johnson,  Meghann and Jamieson,  Jeremy P.},
  year = {2022},
  month = jul,
  pages = {512–520}
}

@misc{https://doi.org/10.48550/arxiv.2007.09845,
  doi = {10.48550/ARXIV.2007.09845},
  url = {https://arxiv.org/abs/2007.09845},
  author = {Woody,  Spencer and Carvalho,  Carlos M. and Hahn,  P. Richard and Murray,  Jared S.},
  title = {Estimating heterogeneous effects of continuous exposures using Bayesian tree ensembles: revisiting the impact of abortion rates on crime},
  publisher = {arXiv},
  year = {2020},
  copyright = {arXiv.org perpetual,  non-exclusive license}
}

@article{Carvalho2010,
  title = {The horseshoe estimator for sparse signals},
  volume = {97},
  ISSN = {1464-3510},
  url = {http://dx.doi.org/10.1093/biomet/asq017},
  DOI = {10.1093/biomet/asq017},
  number = {2},
  journal = {Biometrika},
  publisher = {Oxford University Press (OUP)},
  author = {Carvalho,  C. M. and Polson,  N. G. and Scott,  J. G.},
  year = {2010},
  month = apr,
  pages = {465–480}
}

@article{BrooksGunn1992,
  title = {Effects of early intervention on cognitive function of low birth weight preterm infants},
  volume = {120},
  ISSN = {0022-3476},
  url = {http://dx.doi.org/10.1016/S0022-3476(05)80896-0},
  DOI = {10.1016/s0022-3476(05)80896-0},
  number = {3},
  journal = {The Journal of Pediatrics},
  publisher = {Elsevier BV},
  author = {Brooks-Gunn,  Jeanne and Liaw,  Fong-ruey and Klebanov,  Pamela Kato},
  year = {1992},
  month = mar,
  pages = {350–359}
}

@article{Hill2011,
  title = {Bayesian Nonparametric Modeling for Causal Inference},
  volume = {20},
  ISSN = {1537-2715},
  url = {http://dx.doi.org/10.1198/jcgs.2010.08162},
  DOI = {10.1198/jcgs.2010.08162},
  number = {1},
  journal = {Journal of Computational and Graphical Statistics},
  publisher = {Informa UK Limited},
  author = {Hill,  Jennifer L.},
  year = {2011},
  month = jan,
  pages = {217–240}
}

@article{Chipman2010,
  title = {BART: Bayesian additive regression trees},
  volume = {4},
  ISSN = {1932-6157},
  url = {http://dx.doi.org/10.1214/09-AOAS285},
  DOI = {10.1214/09-aoas285},
  number = {1},
  journal = {The Annals of Applied Statistics},
  publisher = {Institute of Mathematical Statistics},
  author = {Chipman,  Hugh A. and George,  Edward I. and McCulloch,  Robert E.},
  year = {2010},
  month = mar 
}

@inproceedings{Vadillo2010AugmentationIC,
  title={Augmentation in contingency learning under time pressure},
  author={Miguel A. Vadillo},
  year={2010},
  url={https://api.semanticscholar.org/CorpusID:264959031}
}

@inbook{Vallverd2024,
  title = {Do Humans Think Causally,  and How?},
  ISBN = {9789819731879},
  url = {http://dx.doi.org/10.1007/978-981-97-3187-9_4},
  DOI = {10.1007/978-981-97-3187-9_4},
  booktitle = {Causality for Artificial Intelligence},
  publisher = {Springer Nature Singapore},
  author = {Vallverdú,  Jordi},
  year = {2024},
  pages = {33–42}
}

@article{Sloman2015,
  title = {Causality in Thought},
  volume = {66},
  ISSN = {1545-2085},
  url = {http://dx.doi.org/10.1146/annurev-psych-010814-015135},
  DOI = {10.1146/annurev-psych-010814-015135},
  number = {1},
  journal = {Annual Review of Psychology},
  publisher = {Annual Reviews},
  author = {Sloman,  Steven A. and Lagnado,  David},
  year = {2015},
  month = jan,
  pages = {223–247}
}

@article{Penn2007,
  title = {Causal Cognition in Human and Nonhuman Animals: A Comparative,  Critical Review},
  volume = {58},
  ISSN = {1545-2085},
  url = {http://dx.doi.org/10.1146/annurev.psych.58.110405.085555},
  DOI = {10.1146/annurev.psych.58.110405.085555},
  number = {1},
  journal = {Annual Review of Psychology},
  publisher = {Annual Reviews},
  author = {Penn,  Derek C. and Povinelli,  Daniel J.},
  year = {2007},
  month = jan,
  pages = {97–118}
}

@article{Neyman1923,
  title = {On the Application of Probability Theory to Agricultural Experiments.
Essay on Principles. Section 9},
  volume = {5},
  number = {4},
  journal = {Statistical Science},
  author = {Neyman, J.},
  year = {1923},
  pages = {465-480}
}

@article{Neyman1990,
  title = {On the Application of Probability Theory to Agricultural Experiments.
Essay on Principles. Section 9},
  volume = {5},
  number = {4},
  journal = {Statistical Science},
  author = {Neyman, J.},
  year = {1990},
  pages = {465-480}
}

@article{Fisher1935,
  title = {Design of Experiments},
  publisher = {Oliver and Boyd},
  author = {Fisher, R. A.},
  year = {1935},
}

@article{Rubin1974,
  title = {Estimating causal effects of treatments in randomized and nonrandomized studies.},
  volume = {66},
  ISSN = {0022-0663},
  url = {http://dx.doi.org/10.1037/h0037350},
  DOI = {10.1037/h0037350},
  number = {5},
  journal = {Journal of Educational Psychology},
  publisher = {American Psychological Association (APA)},
  author = {Rubin,  Donald B.},
  year = {1974},
  month = oct,
  pages = {688–701}
}

@book{CausalBook,
  title     = "Applied Causal Inference Powered by ML and AI",
  author    = "Chernozhukov, Victor and Hansen, Christian and Kallus, Nathan and Spindler, Martin and Syrgkanis, Vasilis",
  year      = 2024,
  publisher = "Online"
}

@article{Tibshirani1996,
  title = {Regression Shrinkage and Selection Via the Lasso},
  volume = {58},
  ISSN = {1467-9868},
  url = {http://dx.doi.org/10.1111/j.2517-6161.1996.tb02080.x},
  DOI = {10.1111/j.2517-6161.1996.tb02080.x},
  number = {1},
  journal = {Journal of the Royal Statistical Society Series B: Statistical Methodology},
  publisher = {Oxford University Press (OUP)},
  author = {Tibshirani,  Robert},
  year = {1996},
  month = jan,
  pages = {267–288}
}

@misc{https://doi.org/10.48550/arxiv.1712.04912,
  doi = {10.48550/ARXIV.1712.04912},
  url = {https://arxiv.org/abs/1712.04912},
  author = {Nie,  Xinkun and Wager,  Stefan},
  title = {Quasi-Oracle Estimation of Heterogeneous Treatment Effects},
  publisher = {arXiv},
  year = {2017},
  copyright = {arXiv.org perpetual,  non-exclusive license}
}

@article{Cattaneo2018,
  title = {Inference in Linear Regression Models with Many Covariates and Heteroscedasticity},
  volume = {113},
  ISSN = {1537-274X},
  url = {http://dx.doi.org/10.1080/01621459.2017.1328360},
  DOI = {10.1080/01621459.2017.1328360},
  number = {523},
  journal = {Journal of the American Statistical Association},
  publisher = {Informa UK Limited},
  author = {Cattaneo,  Matias D. and Jansson,  Michael and Newey,  Whitney K.},
  year = {2018},
  month = jun,
  pages = {1350–1361}
}

@misc{https://doi.org/10.48550/arxiv.1806.01888,
  doi = {10.48550/ARXIV.1806.01888},
  url = {https://arxiv.org/abs/1806.01888},
  author = {Belloni,  Alexandre and Chernozhukov,  Victor and Chetverikov,  Denis and Hansen,  Christian and Kato,  Kengo},
  title = {High-Dimensional Econometrics and Regularized GMM},
  publisher = {arXiv},
  year = {2018},
  copyright = {arXiv.org perpetual,  non-exclusive license}
}

@article{Zhang2013,
  title = {Confidence Intervals for Low Dimensional Parameters in High Dimensional Linear Models},
  volume = {76},
  ISSN = {1467-9868},
  url = {http://dx.doi.org/10.1111/rssb.12026},
  DOI = {10.1111/rssb.12026},
  number = {1},
  journal = {Journal of the Royal Statistical Society Series B: Statistical Methodology},
  publisher = {Oxford University Press (OUP)},
  author = {Zhang,  Cun-Hui and Zhang,  Stephanie S.},
  year = {2013},
  month = jul,
  pages = {217–242}
}

@article{Belloni2014,
  title = {High-Dimensional Methods and Inference on Structural and Treatment Effects},
  volume = {28},
  ISSN = {0895-3309},
  url = {http://dx.doi.org/10.1257/jep.28.2.29},
  DOI = {10.1257/jep.28.2.29},
  number = {2},
  journal = {Journal of Economic Perspectives},
  publisher = {American Economic Association},
  author = {Belloni,  Alexandre and Chernozhukov,  Victor and Hansen,  Christian},
  year = {2014},
  month = may,
  pages = {29–50}
}

@misc{https://doi.org/10.48550/arxiv.1806.05081,
  doi = {10.48550/ARXIV.1806.05081},
  url = {https://arxiv.org/abs/1806.05081},
  author = {Chernozhukov,  Victor and H\"{a}rdle,  Wolfgang K. and Huang,  Chen and Wang,  Weining},
  title = {LASSO-Driven Inference in Time and Space},
  publisher = {arXiv},
  year = {2018},
  copyright = {arXiv.org perpetual,  non-exclusive license}
}

@misc{https://doi.org/10.48550/arxiv.2403.03240,
  doi = {10.48550/ARXIV.2403.03240},
  url = {https://arxiv.org/abs/2403.03240},
  author = {Kato,  Masahiro},
  title = {Triple/Debiased Lasso for Statistical Inference of Conditional Average Treatment Effects},
  publisher = {arXiv},
  year = {2024},
  copyright = {arXiv.org perpetual,  non-exclusive license}
}

@misc{https://doi.org/10.48550/arxiv.2310.16819,
  doi = {10.48550/ARXIV.2310.16819},
  url = {https://arxiv.org/abs/2310.16819},
  author = {Kato,  Masahiro and Imaizumi,  Masaaki},
  title = {CATE Lasso: Conditional Average Treatment Effect Estimation with High-Dimensional Linear Regression},
  publisher = {arXiv},
  year = {2023},
  copyright = {arXiv.org perpetual,  non-exclusive license}
}

@misc{https://doi.org/10.48550/arxiv.1812.04345,
  doi = {10.48550/ARXIV.1812.04345},
  url = {https://arxiv.org/abs/1812.04345},
  author = {Bach,  Philipp and Chernozhukov,  Victor and Spindler,  Martin},
  title = {Closing the U.S. gender wage gap requires understanding its heterogeneity},
  publisher = {arXiv},
  year = {2018},
  copyright = {arXiv.org perpetual,  non-exclusive license}
}

@article{Chen2019,
  title = {Causal Random Forests Model Using Instrumental Variable Quantile Regression},
  volume = {7},
  ISSN = {2225-1146},
  url = {http://dx.doi.org/10.3390/econometrics7040049},
  DOI = {10.3390/econometrics7040049},
  number = {4},
  journal = {Econometrics},
  publisher = {MDPI AG},
  author = {Chen,  Jau-er and Hsiang,  Chen-Wei},
  year = {2019},
  month = dec,
  pages = {49}
}

@inproceedings{10.1145/3437963.3441722,
author = {Zeng, Shuxi and Bayir, Murat Ali and Pfeiffer, Joseph J. and Charles, Denis and Kiciman, Emre},
title = {Causal Transfer Random Forest: Combining Logged Data and Randomized Experiments for Robust Prediction},
year = {2021},
isbn = {9781450382977},
publisher = {Association for Computing Machinery},
address = {New York, NY, USA},
url = {https://doi.org/10.1145/3437963.3441722},
doi = {10.1145/3437963.3441722},
booktitle = {Proceedings of the 14th ACM International Conference on Web Search and Data Mining},
pages = {211–219},
numpages = {9},
location = {Virtual Event, Israel},
series = {WSDM '21}
}

@article{Ballman2015,
  title = {Biomarker: Predictive or Prognostic?},
  volume = {33},
  ISSN = {1527-7755},
  url = {http://dx.doi.org/10.1200/jco.2015.63.3651},
  DOI = {10.1200/jco.2015.63.3651},
  number = {33},
  journal = {Journal of Clinical Oncology},
  publisher = {American Society of Clinical Oncology (ASCO)},
  author = {Ballman,  Karla V.},
  year = {2015},
  month = nov,
  pages = {3968–3971}
}

@article{ShwartzZiv2022,
  title = {Tabular data: Deep learning is not all you need},
  volume = {81},
  ISSN = {1566-2535},
  url = {http://dx.doi.org/10.1016/j.inffus.2021.11.011},
  DOI = {10.1016/j.inffus.2021.11.011},
  journal = {Information Fusion},
  publisher = {Elsevier BV},
  author = {Shwartz-Ziv,  Ravid and Armon,  Amitai},
  year = {2022},
  month = may,
  pages = {84–90}
}

@ARTICLE{9998482,
  author={Borisov, Vadim and Leemann, Tobias and Seßler, Kathrin and Haug, Johannes and Pawelczyk, Martin and Kasneci, Gjergji},
  journal={IEEE Transactions on Neural Networks and Learning Systems}, 
  title={Deep Neural Networks and Tabular Data: A Survey}, 
  year={2024},
  volume={35},
  number={6},
  pages={7499-7519},
  doi={10.1109/TNNLS.2022.3229161}}

@misc{https://doi.org/10.48550/arxiv.1905.09515,
  doi = {10.48550/ARXIV.1905.09515},
  url = {https://arxiv.org/abs/1905.09515},
  author = {Hahn,  P. Richard and Dorie,  Vincent and Murray,  Jared S.},
  title = {Atlantic Causal Inference Conference (ACIC) Data Analysis Challenge 2017},
  publisher = {arXiv},
  year = {2019},
  copyright = {arXiv.org perpetual,  non-exclusive license}
}

@article{Chesnaye2021,
  title = {An introduction to inverse probability of treatment weighting in observational research},
  volume = {15},
  ISSN = {2048-8513},
  url = {http://dx.doi.org/10.1093/ckj/sfab158},
  DOI = {10.1093/ckj/sfab158},
  number = {1},
  journal = {Clinical Kidney Journal},
  publisher = {Oxford University Press (OUP)},
  author = {Chesnaye,  Nicholas C and Stel,  Vianda S and Tripepi,  Giovanni and Dekker,  Friedo W and Fu,  Edouard L and Zoccali,  Carmine and Jager,  Kitty J},
  year = {2021},
  month = aug,
  pages = {14–20}
}

@article{Dehejia1999,
  title = {Causal Effects in Nonexperimental Studies: Reevaluating the Evaluation of Training Programs},
  volume = {94},
  ISSN = {1537-274X},
  url = {http://dx.doi.org/10.1080/01621459.1999.10473858},
  DOI = {10.1080/01621459.1999.10473858},
  number = {448},
  journal = {Journal of the American Statistical Association},
  publisher = {Informa UK Limited},
  author = {Dehejia,  Rajeev H. and Wahba,  Sadek},
  year = {1999},
  month = dec,
  pages = {1053–1062}
}

@article{Dehejia2002,
  title = {Propensity Score-Matching Methods for Nonexperimental Causal Studies},
  volume = {84},
  ISSN = {1530-9142},
  url = {http://dx.doi.org/10.1162/003465302317331982},
  DOI = {10.1162/003465302317331982},
  number = {1},
  journal = {Review of Economics and Statistics},
  publisher = {MIT Press - Journals},
  author = {Dehejia,  Rajeev H. and Wahba,  Sadek},
  year = {2002},
  month = feb,
  pages = {151–161}
}

@article{Almond2005,
  title = {The Costs of Low Birth Weight},
  volume = {120},
  ISSN = {1531-4650},
  url = {http://dx.doi.org/10.1093/qje/120.3.1031},
  DOI = {10.1093/qje/120.3.1031},
  number = {3},
  journal = {The Quarterly Journal of Economics},
  publisher = {Oxford University Press (OUP)},
  author = {Almond,  D. and Chay,  K. Y. and Lee,  D. S.},
  year = {2005},
  month = aug,
  pages = {1031–1083}
}

@article{Yao2021,
  title = {A Survey on Causal Inference},
  volume = {15},
  ISSN = {1556-472X},
  url = {http://dx.doi.org/10.1145/3444944},
  DOI = {10.1145/3444944},
  number = {5},
  journal = {ACM Transactions on Knowledge Discovery from Data},
  publisher = {Association for Computing Machinery (ACM)},
  author = {Yao,  Liuyi and Chu,  Zhixuan and Li,  Sheng and Li,  Yaliang and Gao,  Jing and Zhang,  Aidong},
  year = {2021},
  month = may,
  pages = {1–46}
}

@InProceedings{pmlr-v222-pitas24a,
  title = 	 {The Fine Print on Tempered Posteriors},
  author =       {Pitas, Konstantinos and Arbel, Julyan},
  booktitle = 	 {Proceedings of the 15th Asian Conference on Machine Learning},
  pages = 	 {1087--1102},
  year = 	 {2024},
  editor = 	 {Yanıkoğlu, Berrin and Buntine, Wray},
  volume = 	 {222},
  series = 	 {Proceedings of Machine Learning Research},
  month = 	 {11--14 Nov},
  publisher =    {PMLR},
  url = 	 {https://proceedings.mlr.press/v222/pitas24a.html}
}

@inproceedings{NEURIPS2022_73e018a0,
 author = {Kapoor, Sanyam and Maddox, Wesley J and Izmailov, Pavel and Wilson, Andrew G},
 booktitle = {Advances in Neural Information Processing Systems},
 editor = {S. Koyejo and S. Mohamed and A. Agarwal and D. Belgrave and K. Cho and A. Oh},
 pages = {18211--18225},
 publisher = {Curran Associates, Inc.},
 title = {On Uncertainty, Tempering, and Data Augmentation in Bayesian Classification},
 url = {https://proceedings.neurips.cc/paper_files/paper/2022/file/73e018a0123b35a3e64269526f9096c9-Paper-Conference.pdf},
 volume = {35},
 year = {2022}
}

@article{https://doi.org/10.48550/arxiv.2008.00029,
  doi = {10.48550/ARXIV.2008.00029},
  url = {https://arxiv.org/abs/2008.00029},
  author = {Adlam,  Ben and Snoek,  Jasper and Smith,  Samuel L.},
  title = {Cold Posteriors and Aleatoric Uncertainty},
  publisher = {arXiv},
  year = {2020},
  copyright = {arXiv.org perpetual,  non-exclusive license}
}

\end{document}